\documentclass{article}

\PassOptionsToPackage{numbers, compress}{natbib}

\usepackage[final]{neurips_2020}

\usepackage{hyperref}       
\usepackage{url}            
\usepackage{booktabs}       
\usepackage{amsfonts}       
\usepackage{nicefrac}       
\usepackage{microtype}      
\usepackage[normalem]{ulem}
\usepackage{multirow}
\usepackage{cancel}
\usepackage{hyperref}
\usepackage{url}
\usepackage{float}
\usepackage{xcolor}
\usepackage[title]{appendix}
\usepackage{wrapfig,lipsum}
\usepackage{mathtools}
\usepackage{multirow}
\usepackage{multicol}
\usepackage{placeins}
\usepackage{algorithm}
\usepackage[algo2e,ruled]{algorithm2e}
\SetKwInput{KwData}{The data}
\SetKwInput{KwResult}{The result}
\usepackage{amsmath}
\usepackage{tikz}
\usetikzlibrary{fit,calc}
\newcommand*{\tikzmk}[1]{\tikz[remember picture,overlay,] \node (#1) {};\ignorespaces}
\newcommand{\boxit}[1]{\tikz[remember picture,overlay]{\node[yshift=3pt,fill=#1,opacity=.25,fit={(A)($(B)+(0\linewidth,0\baselineskip)$)}] {};}\ignorespaces}
\colorlet{pink}{red!40}
\colorlet{blue}{cyan!60}

\usepackage[capitalise]{cleveref}
\Crefformat{figure}{#2{}\scshape{Figure} #1{}#3}
\Crefformat{table}{#2{}\scshape{Table} #1{}#3}

\usepackage{amssymb}

\makeatletter
\algocf@newcommand{KwComment}[1]{%
  \sbox\algocf@inputbox{\hbox{\KwSty{}\algocf@typo}}%
  \ifthenelse{\boolean{algocf@inoutnumbered}}{\relax}{\everypar={\relax}}%
  {\let\\\algocf@newinput\hspace{\wd\algocf@inputbox}\hangindent=\wd\algocf@inputbox\hangafter=\wd\algocf@inputbox\em #1\par}%
  \algocf@linesnumbered
}
\makeatother
\title{Sparse Weight Activation Training}

\author{%
  Md Aamir Raihan, Tor M. Aamodt \\
  Department of Electrical And Computer Engineering\\
  University of British Columbia\\
  Vancouver, BC\\
  \texttt{\{araihan,aamodt\}@ece.ubc.ca} \\
}

\begin{document}

\maketitle
\begin{abstract}
Neural network training is computationally and memory intensive. Sparse training can reduce the burden on emerging hardware platforms designed to accelerate sparse computations, but
it can also affect network convergence. In this work, we propose a novel CNN training algorithm called
Sparse Weight Activation Training (SWAT). SWAT is more computation and memory-efficient than
conventional training. 
SWAT modifies back-propagation based on the empirical insight that convergence during training tends to be robust to the elimination of (i) small magnitude weights during the forward pass and (ii) both small magnitude weights and activations during the backward pass. We evaluate SWAT on recent CNN architectures such as ResNet, VGG,
DenseNet and WideResNet using CIFAR-10, CIFAR-100 and ImageNet datasets.  
For ResNet-50 on ImageNet SWAT reduces total floating-point operations (FLOPs) during training by 80\% 
resulting in a $3.3\times$ training speedup when run on a simulated sparse learning accelerator representative of emerging platforms 
while incurring only 1.63\% reduction in validation accuracy.
Moreover, SWAT reduces memory footprint during the backward pass by 23\% to 50\% for activations and 50\% to 90\% for weights.
Code is available at \url{https://github.com/AamirRaihan/SWAT}.
\end{abstract}

\section{Introduction}

Convolutional Neural Networks (CNNs) are effective at
many complex computer vision tasks including object
recognition~\citep{krizhevsky2012imagenet,szegedy2015going}, object
detection~\citep{NIPS2013_5207, NIPS2015_5638} and image
restoration~\citep{dong2014learning,zhang2017learning}. However, training CNNs
requires significant computation and memory resources.
Software and hardware approaches have been proposed for addressing this challenge.
On the hardware side, graphics processor units (GPUs) are now typically used
for training~\cite{krizhevsky2012imagenet} and recent GPUs from NVIDIA include specialized
Tensor Core hardware specifically to accelerate deep learning~\cite{insideAmpere,Turing,Volta}.
Specialized programmable hardware is being deployed in datacenters by companies such as
Google and Microsoft~\cite{jouppi2017datacenter,chung2018serving}. 
Techniques for reducing computation and memory consumption on existing hardware include 
those reducing the number of training iterations such as
batch normalization~\cite{ioffe2015batch} and enhanced optimization strategies~\cite{kingma2014adam,duchi2011adaptive}
and those reducing computations per iteration.  
Examples of the latter, which may be effective with appropriate hardware support,
include techniques such as
quantization~\citep{zhou2016dorefa,choi2018bridging,wu2016quantized,wang2018training}, use of fixed-point
instead of floating-point~\citep{wu2018training,das2018mixed}, sparsification~\cite{wei2017minimal} and dimensionality reduction~\cite{liu2018dynamic}. 
This paper introduces {\em Sparse Weight Activation Training} (SWAT), which significantly extends the sparsification approach.

Training involves repeated application of forward and backward passes.  Prior
research on introducing sparsity during training has focused on sparsifying the backward
pass.  While model compression~\cite{han2015deep,molchanov2017variational,louizos2017learning,liu2017learning,li2016pruning,he2017channel,luo2017thinet,wen2016learning,molchanov2016pruning} introduces sparsification into the forward pass, it typically does
so by introducing additional training phases which increase overall training time.
Amdahl's Law~\cite{amdahl1967validity} implies overall speedup is limited by 
the fraction of original execution time spent on computations that are not sped up 
by system changes.  To reduce training time significantly by reducing computations per training
iteration it is necessary to address both forward and backward passes.
SWAT 
introduces sparsification into both forward and backward passes 
and is suitable for emerging hardware platforms 
containing support for sparse matrix operations. Such hardware is now available. 
For example recently announced Ampere GPU architecture~\cite{insideAmpere} 
includes support for exploiting sparsity.
In addition, there is a growing body of research on hardware accelerators for sparse networks~\cite{parashar2017scnn,delmas2019bit,zhang2016cambricon,albericio2016cnvlutin}
and we demonstrate, via hardware simulation, that SWAT can potentially train $5.9\times$ faster when such accelerators become available.

While SWAT employs sparsity it does so with the objective of {\em reducing training time {\bf not} performing model compression}.
The contributions of this paper are:
\begin{itemize}
\item An empirical sensitivity analysis of approaches to inducing sparsity in network training;
\item SWAT, a training algorithm that introduces sparsity in weights and activations resulting
  in reduced execution time in both forward and backward passes of training;
\item An empirical evaluation showing SWAT is effective on complex models and datasets. 
\end{itemize}

\section{Related work}
Below we summarize the works most closely related to SWAT.

\paragraph{Network pruning:}
\citet{lecun1990optimal} proposed removing network parameters 
using second-order information of the loss function to improve generalization 
and reduce training and inference time.  
More recently, pruning has been explored primarily as a way to improve the efficiency and storage requirements of
inference but at the expense of increasing training time, contrary to the objective of this paper.  Specifically, 
\citet{han2015deep} showed how to substantially reduce network parameters 
while improving validation accuracy by pruning weights based upon their magnitude
combined with a subsequent retraining phase that fine-tunes the remaining weights.
Regularization techniques can be employed to learn a pruned network~\cite{molchanov2017variational, louizos2017learning, golub2018full}.
Various approaches to structured pruning~\cite{wen2016learning,molchanov2016pruning,li2016pruning,liu2017learning,he2017channel,luo2017thinet},
ensure entire channels or filters are removed to reduce inference execution time on the vector hardware found in GPUs.
\paragraph{Reducing per-iteration training overhead:} 
MeProp~\citep{sun2017meprop, wei2017minimal} reduces computations
during training by back propagating only the
highest magnitude gradient component and setting other component to zero.
As shown in Section~\ref{sec:sensitivity}, on complex networks and models training is
very sensitive to the fraction of 
gradient components set to zero.
\citet{liu2018dynamic} proposed reducing computation during training and inference by constructing a
dynamic sparse graph~(DSG) using random projection.   
DSG incurs accuracy loss of $3\%$ on ImageNet at a sparsity of $50\%$.
\citet{negar2020} propose to reduce backward pass computations by 
performing convolutions only on gradient components that change
substantially from the prior iteration. 
They reduced the overall computation in the backward pass by up to 90\% with minimum loss in accuracy. Their approach of backward pass sparsification is orthogonal to SWAT. 

\paragraph{Sparse Learning:} 
Sparse learning attempts to learn a sparse representation during training, generally as a way to achieve model compression and reduce
computation during inference.  Since sparse learning introduces sparsity during training it can potentially reduce training time
but pruning weights by itself does not reduce weight gradient computation (Equation~\ref{eqn:3}).
Many sparse learning algorithms start with a random sparse network, then repeat a cycle of training, pruning and regrowth.  
Sparse Evolutionary Training~(SET)~\cite{mocanu2018scalable} prunes the most negative and smallest positive weights then randomly selects latent (i.e., missing) weights for regrowth.
Dynamic Sparse Reparameterization~(DSR)~\cite{mostafa2019parameter} uses a global adaptive threshold for pruning 
and randomly regrows latent weights in a layer proportionally to the number of active (non-zero) weights
in that same layer. 
Sparse Network From Scratch~(SNFS)~\cite{dettmers2019sparse} 
further improves performance
using magnitude-based pruning and momentum for determining the regrowth across layers. 
Rigging the Lottery Ticket~(RigL)~\cite{evci2019rigging} uses an instantaneous gradient as its regrowth criteria. 
Dynamic Sparse Training~(DST)~\cite{liu2020dynamic} 
defines a trainable mask to determine which weights to prune.
\textcolor{black}{Recently \citet{kusupati2020soft} proposes a novel state-of-the-art method of finding per layer learnable threshold which reduces 
the FLOPs during inference by employing a non-unform sparsity budget across layers. } 

\begin{table}[t]
\centering
\caption{Comparison of SWAT with related works}
\label{table-1:comparision}
\resizebox{\textwidth}{!}{%
\begin{tabular}{c|c|c|c|c|c|c}
\toprule
\multirow{3}{*}{Algorithms} & \multirow{3}{*}{\begin{tabular}[c]{@{}c@{}}Sparsity\\ Across\\ Layer\end{tabular}} & \multirow{3}{*}{\begin{tabular}[c]{@{}c@{}}Sparse\\ Forward \\ Pass\end{tabular}} & \multicolumn{2}{c|}{Sparse Backward Pass} & \multirow{3}{*}{\begin{tabular}[c]{@{}c@{}}Unstructured \\ Sparse \\ Network\end{tabular}} & \multirow{3}{*}{\begin{tabular}[c]{@{}c@{}}Structured \\ Sparse\\ Network\end{tabular}} \\ \cline{4-5}
 &  &  & \multirow{2}{*}{Input Gradient} & \multirow{2}{*}{Weight Gradient} &  &  \\
 &  &  &  &  &  &  \\ 
	\midrule
\begin{tabular}[c]{@{}c@{}}Network\\ Pruning\end{tabular} & \begin{tabular}[c]{@{}c@{}}Fixed/\\ Variable\end{tabular} & \begin{tabular}[c]{@{}c@{}}Yes/\\ Gradual\end{tabular} & \begin{tabular}[c]{@{}c@{}}Yes/\\ Gradual\end{tabular} & No & \multicolumn{2}{c}{Depend on algorithm} \\ 
meProp~\cite{sun2017meprop} & Fixed & No & Yes & Yes & No & - \\ 
DSG~\cite{liu2018dynamic} & Fixed & Yes & Yes & Yes & Yes & - \\ 
SET~\cite{mocanu2018scalable} & Fixed & Yes & Yes & No & Yes & - \\ 
DSR~\cite{mostafa2019parameter} & Variable & Yes & Yes & No & Yes & - \\ 
SNFS~\cite{dettmers2019sparse} & Variable & Yes & Yes & No & Yes & - \\ 
RigL~\cite{evci2019rigging} & Fixed & Yes & Yes & No & Yes & - \\ \hline
SWAT& \begin{tabular}[c]{@{}c@{}}Fixed/\\ Variable\end{tabular} & Yes & Yes & Yes & Yes & Yes \\ 
\bottomrule
\end{tabular}%
}
\end{table}

In contrast, SWAT employs a unified training phase where the algorithm continuously explores sparse topologies during training 
by employing simple magnitude based thresholds to determine which weight and activations components to operate upon.
In addition, the sparsifying 
function used in SWAT can be adapted to induce structured sparse topologies 
and varying sparsity across layers. 
\autoref{table-1:comparision} summarizes the differences between SWAT and recent related work on increasing sparsity.
\textcolor{black}{The column ``Sparsity Across Layer" indicates whether the layer sparsity is constant during training. The columns labeled 
``Input Gradient" and ``Weight Gradient" represent whether the input gradient or weight gradient computation is sparse during training.}

\section{Sparse weight activation training}
We begin with preliminaries, describe a sensitivity study motivating SWAT then describe SWAT and several enhancements.
\subsection{Preliminaries}
We consider a CNN 
trained using mini-batch stochastic gradient descent, where the $l^{th}$ layer maps input activations $a_{l-1}$ to outputs $a_l$ using function $f_{l}$:
\begin{equation}\label{eq:forw1}
  a_{l}=f_{l}(a_{l-1},w_{l})
\end{equation}
where $w_l$ are layer $l$'s weights.
During back-propagation the $l^{th}$ layer receives the gradient of the loss 
with respect to its output activation ($\bigtriangledown_{a_{l}}$). 
This is used to compute the gradient of the loss with respect to its input activation ($\bigtriangledown_{a_{l-1}}$) and weight ($\bigtriangledown_{w_{l}}$) using function $G_{l}$ and $H_{l}$ respectively. 
Thus, the backward pass for the $l^{th}$ layer can be defined as:
\begin{align}
  &\bigtriangledown_{a_{l-1}}= G_{l}(\bigtriangledown_{a_{l}},w_{l}), \label{eqn:2} \\
  &\bigtriangledown_{w_{l}} = H_{l}(\bigtriangledown_{a_{l}},a_{l-1}) \label{eqn:3}
\end{align}
We induce sparsity on a tensor
by retaining the values for the $K$ highest magnitude elements
and setting the remaining elements to zero.
Building on the notion of a ``Top-k'' query in databases~\cite{shanbhag2018efficient}
and simlar to~\citet{sun2017meprop} we call this process Top-$K$ sparsification.

\subsection{Sensitivity Analysis}
\label{sec:sensitivity}
We begin by studying the sensitivity of network convergence
by applying Top-$K$ sparsification to weights~($w_{l}$), activations~($a_{l-1}$) and/or back-propagated error gradients~($\bigtriangledown_{a_{l}}$) during training.
For these experiments, which help motivate SWAT, 
we evaluate DenseNet-121, VGG-16 and ResNet-18 on the CIFAR-100 dataset. 
We run each experiment three times and report the mean value.

\begin{figure}[H]
\centering
\noindent\begin{minipage}{.5\linewidth}
\centering
\includegraphics[scale=0.35]{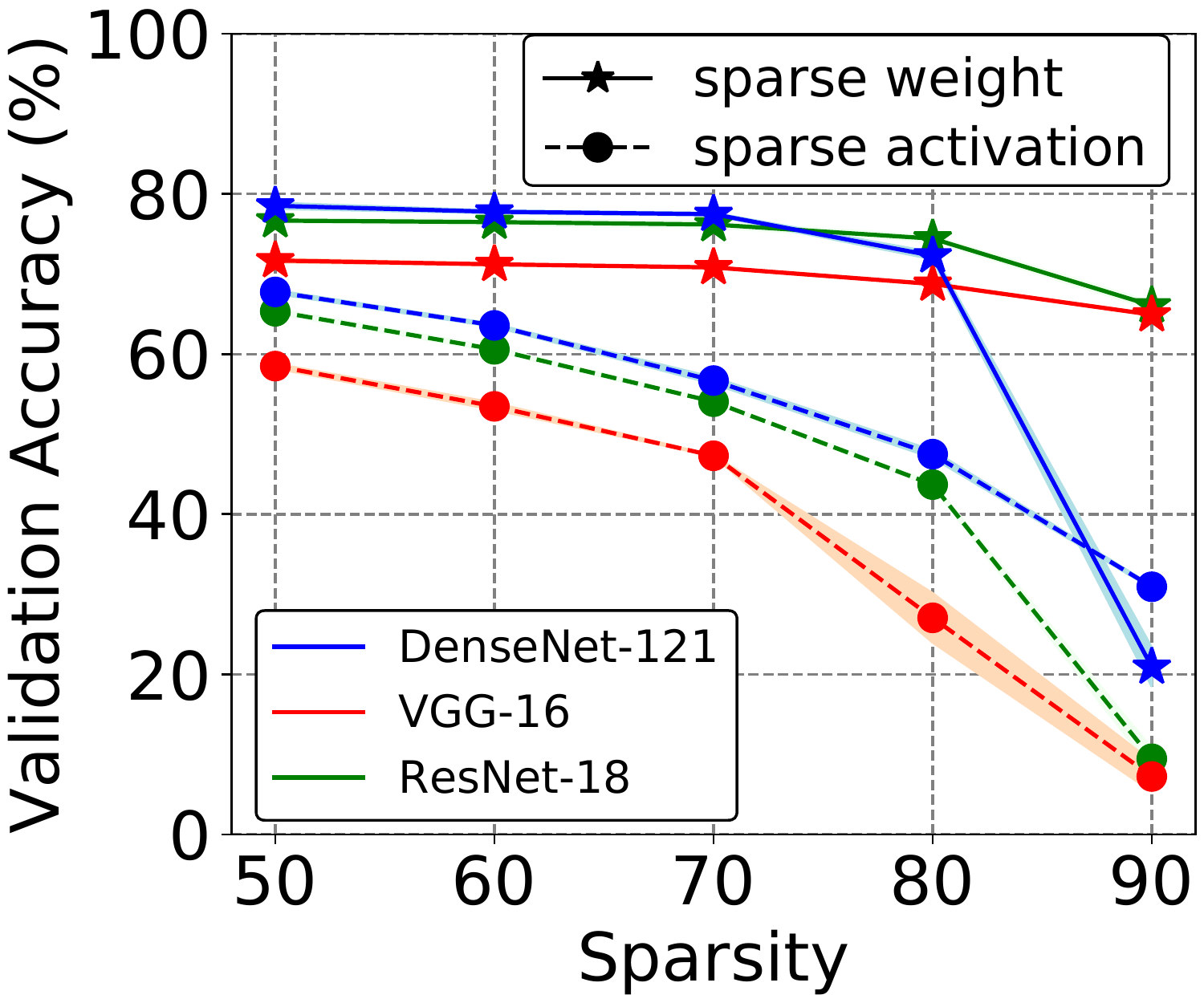}
\caption{Forward Pass Sensitivity Analysis.}
  \label{fig:SensitivityForwardPass}        	
\end{minipage}%
\begin{minipage}{.5\linewidth}
\centering
\includegraphics[scale=0.35]{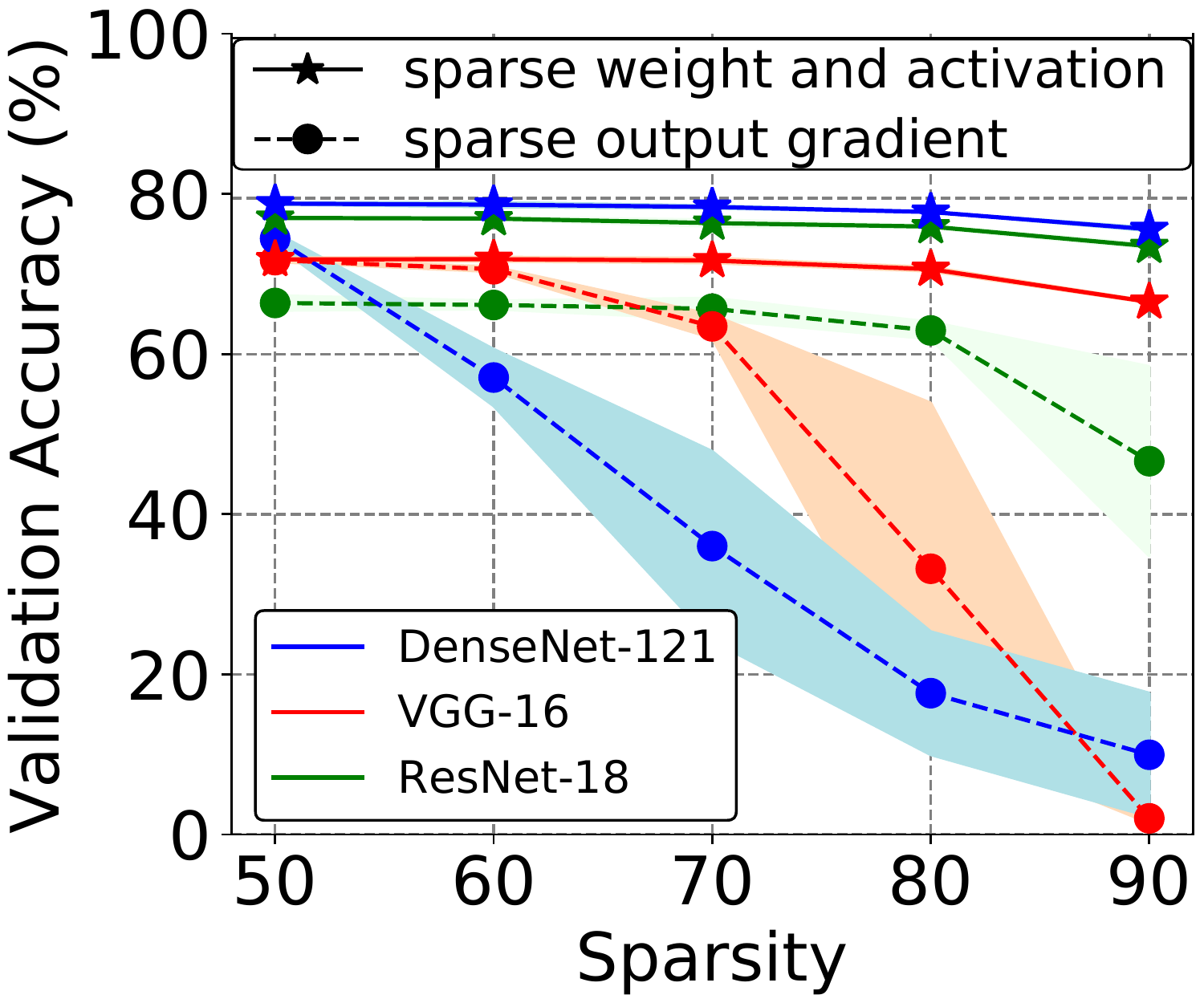}
\caption{Backward Pass Sensitivity Analysis.} 
\label{fig:SensitivityBackwardPass}        	
\end{minipage}
\end{figure}

\autoref{fig:SensitivityForwardPass} plots the impact on validation accuracy of applying varying
degrees of Top-$K$ sparsification to weights or activations during computations in the forward pass (Equation~\ref{eq:forw1}).
In this experiment, when applying top-$K$ sparsification in the forward pass weights or activations are not permanently removed. 
Rather, low magnitude weights or activations are removed temporarily during the forward pass and restored before the backward pass.
Thus, in this experiment the backward pass uses unmodified weights and activations without applying any sparsification
and all weights are updated by the resulting dense weight gradient.
The data in \autoref{fig:SensitivityForwardPass} suggests convergence is more robust to Top-$K$ sparsification 
of weights versus activations during the forward pass.
 \textcolor{black}{The data shows that inducing high activation sparsity hurts accuracy after a certain point, confirming similar observations by \citet{georgiadis2019accelerating} and \citet{kurtzinducing}.}

\autoref{fig:SensitivityBackwardPass} plots the impact on validation accuracy of applying varying
degrees of Top-$K$ sparsification to both weights and activations (labeled ``sparse weight and activation'')
or to only output gradients (labeled ``sparse output gradient'')
during computations in the backward pass (Equations~\ref{eqn:2} and~\ref{eqn:3}). 
\textcolor{black}{``Weights and activations" or ``back-propagated output error gradients" are sparsified before performing the convolutions to generate weight and input gradients.  The generated gradients are dense since a convolution between a sparse and dense inputs will in general produce a dense output. The resulting dense weight gradients are used in the parameter update stage. We note that this process differs, for example, from recent approaches to sparsifying inference (e.g.,~\cite{kusupati2020soft,zhu2017prune,evci2019rigging}) which employ dense back-propagated output gradients during convolution to generate weight gradients that are masked during the parameter update stage. }
The ``sparse weight and activation'' curve shows that convergence is relatively insensitive to applying Top-$K$ sparsification.
In contrast, the ``sparse output gradient'' curve
shows that convergence is sensitive to applying Top-$K$ sparsification to back-propagated error gradients ($\bigtriangledown_{a_{l}}$).
The latter observation indicates that meProp, which drops back-propagated error-gradients, will suffer convergence issues
on larger networks.

\subsection{The SWAT Algorithm} 
The analysis above suggests two strategies: 
In the forward pass use sparse weights (but {\em not} activations) and in the backward pass use 
sparse weights and activations (but {\em not} gradients).

Sparse weight activation training (SWAT) embodies these two strategies as follows (for pseudo-code see supplementary material):
During each training pass (forward and backward iteration on a mini-batch)
a sparse weight topology is induced during the forward pass using the Top-$K$ function.  
This partitions weights into active~(i.e., Top-K) and non-active sets for the current iteration. 
The forward pass uses the active weights. In the backward pass, 
the full gradients and active weights 
are used in Equation~\ref{eqn:2} 
and the full gradients and highly activated neurons (Top-$K$ sparsified activations) are used in Equation~\ref{eqn:3}.
The later generate dense weight gradients that are used to update {\em both} active and non-active weights.
The updates to non-active weights mean the topology can change from iteration to iteration. 
\textcolor{black}{This enables SWAT to perform dynamic topology exploration: Backpropagation with sparse weights and activations approximates backpropagation on a network with sparse connectivity and sparsely activated neurons. The dense gradients generated during back-propagation minimize loss for the current sparse connectivity. However, the updated weights resulting from the dense gradients will potentially lead to a new sparse network since non-active weights are also updated. This process captures fine-grained temporal importance of connectivity during training. Section~\ref{sec:ablation} shows quantitatively, the importance of unmasked gradient updates and dynamic exploration of connectivity. }

\subsubsection{Top-K Channel Selection}\label{topkchannelselection} 

\begin{wrapfigure}{r}{3.5cm}
  \vspace{-0.3in}
\includegraphics[width=3.5cm]{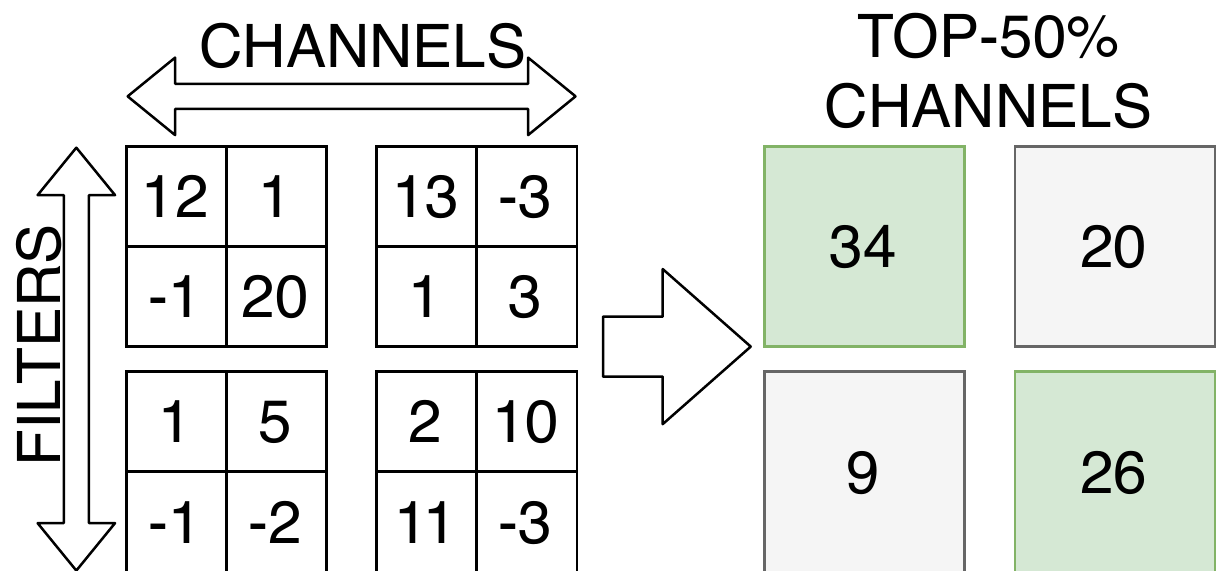}
  \vspace{-0.2in}
\caption{Top-K Channel Selection.}
\label{fig:Top-K-channel-filters}
\end{wrapfigure}

Top-$K$ sparsification induces fine-grained sparsity.
\textcolor{black}{SWAT can instead induce structured sparsity on weights by pruning channels.
Similarly to \cite{cai2020onceforall,liu2017learning,mao2017exploring,li2016pruning}, the saliency criteria for selecting channels is $L_1$ norm.} 
\autoref{fig:Top-K-channel-filters} illustrates 
using channel $L_{1}$ norm to select 50\% of channel (Top-50\%).
The squares on the right side contain the channel $L_1$ norm and
lower $L_1$ norm channels are set as non-active (lightly shaded).
\textcolor{black}{The importance of channel’s is consider independently, i.e., different filters can select different active channels.}

\subsubsection{Sparsity Distribution}\label{sparsitydistribution}
The objective of SWAT is to reduce training time while maintaining accuracy.
So far we have assumed the per layer sparsity of weights and activations is equal to the target sparsity.
Prior work~\cite{evci2019rigging, dettmers2019sparse} has demonstrated that a non-uniform distribution of active weights and 
activation sparsity can improve accuracy for a given sparsity. 
We explore three strategies to distributing sparsity.  All three of the following variants 
employ magnitude comparisons to select which tensor elements to set to zero to induce sparsity.
We say an element of a tensor is unmasked if it is not forced to zero before being used in a computation.
For all three techniques below the fraction of unmasked elements in a given layer is identical for weights and activations.

\paragraph{Uniform~(SWAT-U):} 
Similar to others (e.g., \citet{evci2019rigging}) we found that keeping first layer dense improves validation accuracy.
For $\text{SWAT-U}$ we keep the first layer dense and apply the same sparsity threshold uniformly across all other layers.

\paragraph{Erdos-Renyi-Kernel~(SWAT-ERK):} 
For SWAT-ERK active weights and unmasked activations are distributed across layers 
by taking into account layer dimensions and setting a per layer threshold for magnitude based pruning.
Following \citet{evci2019rigging} higher sparsity is allocated to layers containing more parameters.

\paragraph{Momentum~(SWAT-M):} 
For SWAT-M active weights and unmasked activations are distributed across layers 
such that less non-zero elements are retained in layers with smaller average momentum for active weights.
This approach is inspired by \citet{dettmers2019sparse}. 

\begin{figure}[H]
\centering
\noindent\begin{minipage}{0.38\linewidth}
\centering
\includegraphics[scale=0.42]{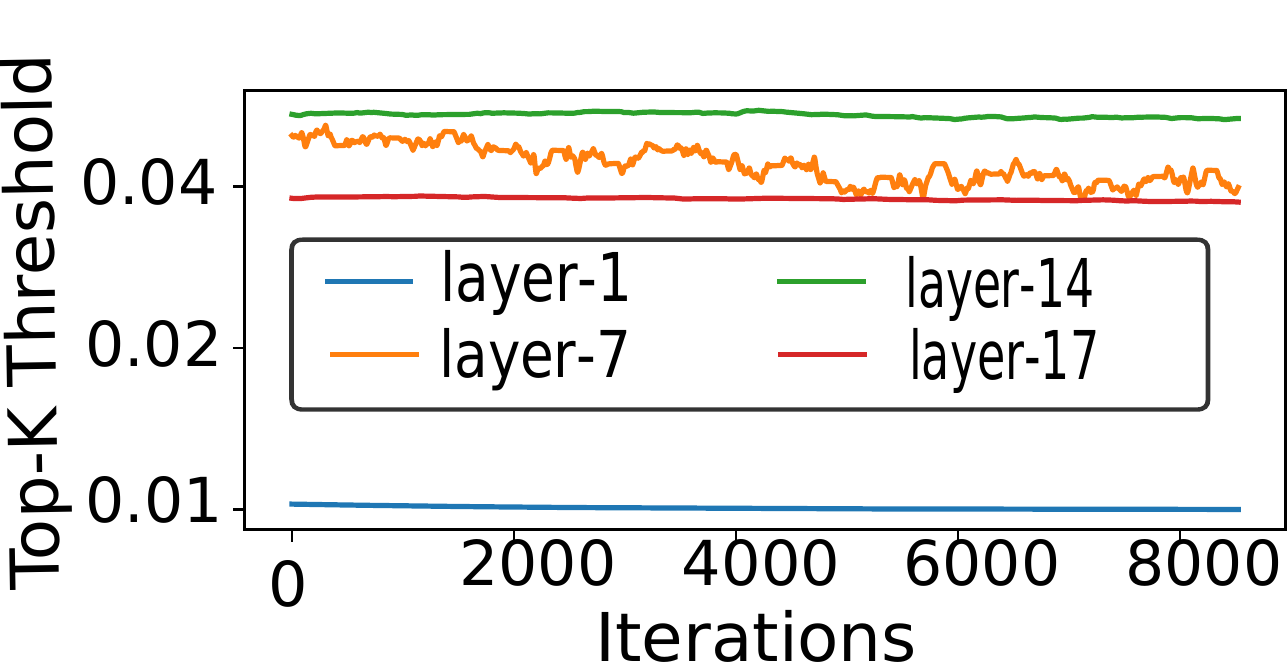}
\caption{Top-$K$ weight threshold versus training iteration}
\label{fig:efficienttopkimplementation}
\end{minipage}%
\hfill
\begin{minipage}{.58\linewidth}
\centering
\includegraphics[scale=0.26]{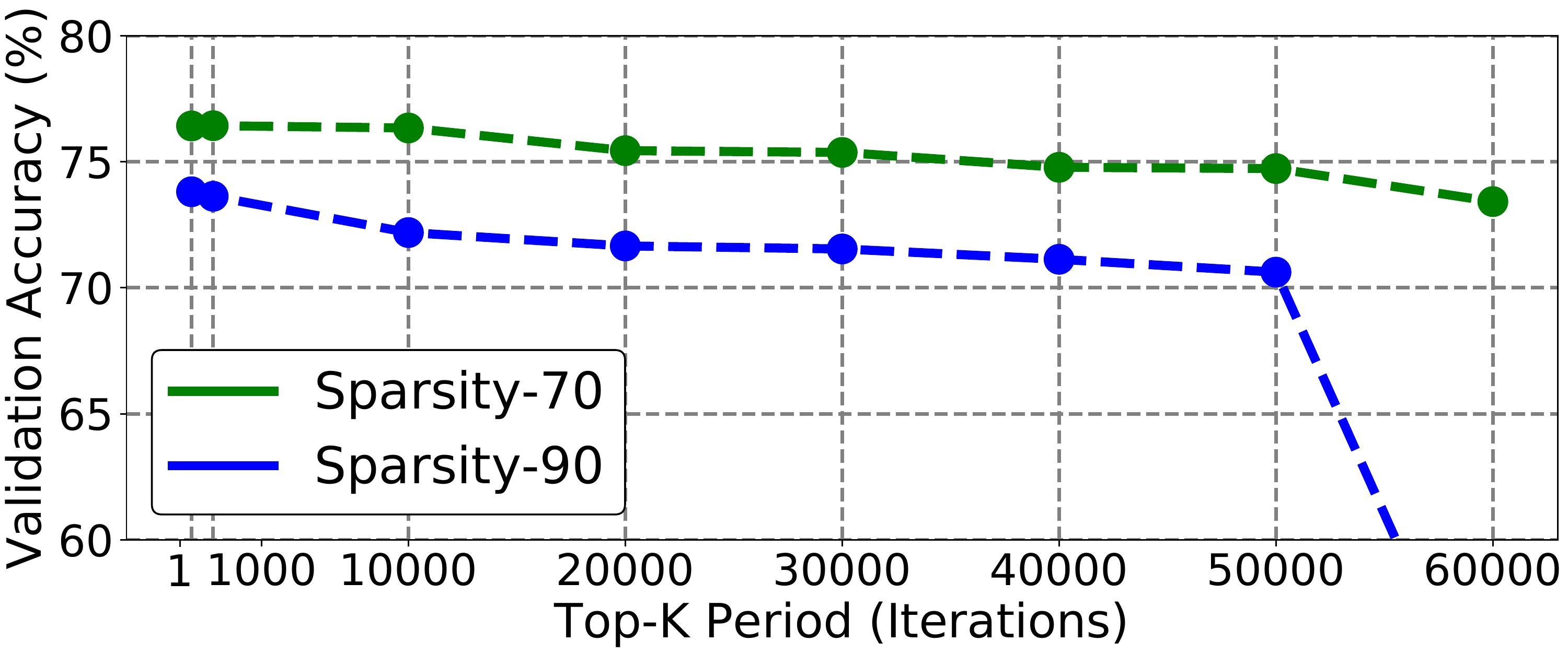}
\caption{Impact of Top-$K$ sampling period.}
\label{fig:efficientTopK}
\end{minipage}
\end{figure}

\subsubsection{Efficient Top-K Threshold Calculation}\label{efficienttopkimplementation} 
The variants of SWAT described above induce sparsity by examining the magnitude of tensor elements
and this incurs some overhead.  Naively, the Top-$K$ operation could be 
performed on a 1-dimensional array of size $N$ in $O(N\log{}N)$ using sorting. 
The overhead can be reduced to $O(N)$ using a threshold operation where elements
with magnitude greater than a threshold are retained an others are treated as zeros.
The $K$-th largest element can be found in $O(N)$ average time complexity using Quickselect~\citep{Hoare:1961:AF:366622.366647} or
in $\theta(N)$ time using either BFPRT~\citep{BLUM1973448} or Introselect~\citep{introselect} and
efficient parallel implementations of Top-$K$ have been proposed for GPUs~\cite{shanbhag2018efficient}.

\autoref{fig:efficienttopkimplementation} plots the threshold value required to achieve 90\% weight sparsity 
versus training iteration for SWAT-U and unstructured pruning four representative layers of ResNet-18 while training using CIFAR-100.
This data suggests that for a given layer the magnitude of the $K$-th largest element is almost constant during training.
Thus, we explored whether we can set a per layer threshold for weights and activations on the first iteration and only
update it periodically.

\autoref{fig:efficientTopK} plots the final Top-1 validation accuracy 
after training converges versus Top-$K$ sampling period for
SWAT-U and unstructured pruning applied to ResNet-18 trained using CIFAR 100.
Here 391 iterations (x-axis) corresponds to a single epoch.
The data indicates that for sampling intervals up to 1000 iterations validation accuracy is not degraded significantly.

\section{Experiments}
\label{result}
Below we present results for validation accuracy, theoretical reduction in floating-point operations (FLOPs) 
during training and estimates of training speedup on a simulated sparse accelerator.  
While {\bf not} our primary objective we also report theoretical FLOPs reduction for inference.

\subsection{Methodology} 
\label{id}

We measure validation accuracy of SWAT by implementing custom convolution and
linear layers in PyTorch 1.1.0~\citep{paszke2017automatic}.  Inside each custom
PyTorch layer we perform sparsification before performing the layer forward or
backward pass computation.  To obtain accuracy measurements in a reasonable time these
 custom layers invoke NVIDIA's cuDNN library using Pytorch's C++ interface.

We estimate {\em potential} for training time reduction using an analytical
model to measure total floating-point operations and an architecture simulator
modeling a sparse DNN training accelerator based upon an extension of the
Bit-Tactical inference accelerator~\cite{delmas2019bit}.  

We employ standard training schedules for training ResNet~\citep{he2016deep},
VGG~\citep{simonyan2014very} with batch-normalization~\citep{ioffe2015batch},
DenseNet~\citep{huang2017densely} and Wide Residual
Network~\citep{zagoruyko2016wide} on CIFAR10 and
CIFAR100~\citep{krizhevsky2009learning}. We use SGD with momentum as an
optimization algorithm with an initial learning rate of $0.1$, momentum of
$0.9$, and weight decay $\lambda$ of $0.0005$.  For training runs with ImageNet
we employ the augmentation technique proposed by
\citet{krizhevsky2012imagenet}: $224\times 224$ random crops from the input
images or their horizontal flip are used for training. Networks are trained
with label smoothing~\cite{szegedy2016rethinking} of $0.1$ for $90$ epochs with
a batch size of 256 samples on a system with eight NVIDIA 2080Ti GPUs.  The
learning rate schedule starts with a linear warm-up reaching its maximum of 0.1
at epoch 5 and is reduced by $(1/10)$ at epochs $30^{th}$, $60^{th}$ and
$80^{th}$. The optimization method is SGD with Nesterov momentum of $0.9$ and
weight decay $\lambda$ of $0.0001$.  Results for ImageNet use a Top-$K$ threshold 
recomputed every 1000 iterations while those for CIFAR-10 recompute the Top-$K$ threshold
every iteration.  Due to time and resource constraints below we
report SWAT-ERK and SWAT-M results only for CIFAR-10 and unstructured sparsity.

The supplementary material includes detailed hyperparameters, results for CIFAR-100,
ablation study of the accuracy impact of performing Top-$K$ on different subsets 
of weight and activation tensors.

\subsection{Unstructured SWAT}
\paragraph{CIFAR-10:}

\begin{table}[]
	\caption{Unstructured SWAT on the CIFAR-10 dataset. \textbf{W:} Weight, \textbf{Act:} Activation, \textbf{BA:} Baseline Accuracy, \textbf{AC:} Accuracy Change, \textbf{MC:} Model Compression, \textbf{DS:} Default Sparsity.}
\label{table:comparisioncifar10unstructured}
\resizebox{\textwidth}{!}{%
\begin{tabular}{cccccccccc}
\hline
\multirow{3}{*}{Network} & \multirow{3}{*}{Methods} & \multicolumn{2}{c}{Training Sparsity} & \multicolumn{2}{c}{Top-1} & \multicolumn{2}{c}{\bf Training} & \multicolumn{2}{c}{Inference} \\ \cline{3-10} 
 &  & \multirow{2}{*}{\begin{tabular}[c]{@{}c@{}}W\\  (\%)\end{tabular}} & \multirow{2}{*}{\begin{tabular}[c]{@{}c@{}}Act\\ (\%)\end{tabular}} & \multirow{2}{*}{\begin{tabular}[c]{@{}c@{}}Acc. $\pm$ SD ($\sigma$)  \\ (\%)\end{tabular}} & \multirow{2}{*}{\begin{tabular}[c]{@{}c@{}}BA /\\AC\end{tabular}} & \multirow{2}{*}{\begin{tabular}[c]{@{}c@{}}\bf FLOPs $\downarrow$\\  (\%)\end{tabular}} & \multirow{2}{*}{\begin{tabular}[c]{@{}c@{}}Act $\downarrow$\\  (\%)\end{tabular}} & \multirow{2}{*}{\begin{tabular}[c]{@{}c@{}}MC \\ ($\times$)\end{tabular}} & \multirow{2}{*}{\begin{tabular}[c]{@{}c@{}}FLOPs $\downarrow$\\  (\%)\end{tabular}} \\
 &  &  &  &  &  &  &  &  &  \\ \hline
\multirow{5}{*}{VGG-16} & SNFS~\cite{dettmers2019sparse} & 95.0 & DS & 93.31 & 93.41 / -0.10 & 57.0 & DS & 20.0 & 62.9 \\
 & DST~\cite{liu2020dynamic} & 96.2 & DS & 93.02 & 93.75 / -0.73 & 75.7 & DS & 26.3 & 83.2 \\
& SWAT-U & 90.0 & 90.0 & 91.95 $\pm$ 0.06 & 93.30 / -1.35 & 89.7 & 36.2 & 10.0 & 89.5 \\
 & SWAT-ERK & 95.0 & 82.0 & 92.50 $\pm$0.07 & 93.30 / -0.80 & 89.5 & 33.0 & 20.0 & 89.5 \\
 & SWAT-M & 95.0 & 65.0 & \textbf{93.41 $\pm$0.05} & 93.30 / \textbf{+0.11} & 64.0 & 25.0 & 20.0 & 58.4 \\ \hline
\multirow{5}{*}{WRN-16-8} & SNFS~\cite{dettmers2019sparse} & 95.0 & DS & 94.38 & 95.43 / -1.05 & 81.8 & DS & 20.0 & 88.0 \\
 & DST~\cite{liu2020dynamic} & 95.4 & DS & 94.73 & 95.18 / -0.45 & 83.3 & DS & 21.7 & 91.4 \\
 & SWAT-U & 90.0 & 90.0 & \textbf{95.13 $\pm$0.11} & 95.10 / \textbf{+0.03} & 90.0 & 49.0 & 10.0 & 90.0 \\
 & SWAT-ERK & 95.0 & 84.0 & 95.00 $\pm$0.12 & 95.10 / -0.10 & 91.4 & 45.8 & 20.0 & 91.7 \\
 & SWAT-M & 95.0 & 78.0 & 94.97 $\pm$0.04 & 95.10 / -0.13 & 86.3 & 42.5 & 20.0 & 85.9 \\ \hline
\multirow{3}{*}{DenseNet-121} & SWAT-U & 90.0 & 90.0 & \textbf{94.48 $\pm$0.06} & 94.46 / \textbf{+0.02} & 89.8 & 44.0 & 10.0 & 89.8 \\
 & SWAT-ERK & 90.0 & 88.0 & 94.14 $\pm$0.11 & 94.46 / -0.32 & 89.7 & 43.0 & 10.0 & 89.6 \\
 & SWAT-M & 90.0 & 86.0 & 94.29 $\pm$0.11 & 94.46 / -0.17 & 84.2 & 42.0 & 10.0 & 83.6 \\ \hline
\end{tabular}%
}
\end{table}
\begin{table}[]
	\caption{Unstructured SWAT on the ImageNet dataset. \textbf{W:} Weight, \textbf{Act:} Activation, \textbf{BA:} Baseline Accuracy , \textbf{AC:} Accuracy Change, \textbf{MC:} Model Compression, \textbf{DS:} Default Sparsity.}
\label{table:comparisionimagenetunstructured}
\resizebox{\textwidth}{!}{%
\begin{tabular}{cccccccccc}
\hline
\multirow{3}{*}{Network} & \multirow{3}{*}{Methods} & \multicolumn{2}{c}{\multirow{2}{*}{Training Sparsity}} & \multicolumn{2}{c}{\multirow{2}{*}{Top-1}} & \multicolumn{2}{c}{\multirow{2}{*}{\bf Training}} & \multicolumn{2}{c}{\multirow{2}{*}{Inference}} \\
 &  & \multicolumn{2}{c}{} & \multicolumn{2}{c}{} & \multicolumn{2}{c}{} & \multicolumn{2}{c}{} \\ \cline{3-10} 
 &  & W (\%) & Act (\%) & \begin{tabular}[c]{@{}c@{}}Acc. $\pm$ SD ($\sigma$)\\ (\%)\end{tabular} & \begin{tabular}[c]{@{}c@{}}BA / \\ AC\end{tabular} & \begin{tabular}[c]{@{}c@{}}\bf FLOPs $\downarrow$\\  (\%)\end{tabular} & \begin{tabular}[c]{@{}c@{}}Act $\downarrow$\\  (\%)\end{tabular} & \begin{tabular}[c]{@{}c@{}}MC\\  ($\times$)\end{tabular} & \begin{tabular}[c]{@{}c@{}}FLOPs $\downarrow$\\  (\%)\end{tabular} \\ \hline
\multirow{12}{*}{ResNet-50} & \multirow{2}{*}{SET~\cite{mocanu2018scalable}} & 80.0 & \multirow{2}{*}{\begin{tabular}[c]{@{}c@{}}DS\end{tabular}} & 73.4 $\pm$0.32 & 76.8 / -3.4 & 58.1 & DS & 3.4 & 73.0 \\
 &  & 90.0 &  & 71.3 $\pm$0.24 & 76.8 / -5.5 & 63.8 & DS & 5.0 & 82.1 \\ \cline{2-10} 
 & \multirow{2}{*}{DSR~\cite{mostafa2019parameter}} & 80.0 & \multirow{2}{*}{DS} & 74.1 $\pm$0.17 & 76.8 / -2.7 & 51.6 & DS & 3.4 & 59.4 \\
 &  & 90.0 &  & 71.9 $\pm$0.07 & 76.8 / -4.9 & 58.9 & DS & 5.0 & 70.7 \\ \cline{2-10} 
 & \multirow{2}{*}{SNFS~\cite{dettmers2019sparse}} & 80.0 & \multirow{2}{*}{DS} & 74.9 $\pm$0.07 & 77.0 / -2.1 & 45.8 & DS & 5.0 & 43.3 \\
 &  & 90.0 &  & 72.9 $\pm$0.07 & 77.0 / -4.1 & 57.6 & DS & 10.0 & 59.7 \\ \cline{2-10} 
 & \multirow{2}{*}{RigL~\cite{evci2019rigging}} & 80.0 & \multirow{2}{*}{DS} & 74.6 $\pm$0.06 & 76.8 / -2.2 & 67.2 & DS & 5.0 & 77.7 \\
 &  & 90.0 &  & 72.0 $\pm$0.05 & 76.8 / -4.8 & 74.1 & DS & 10.0 & 87.4 \\ \cline{2-10} 
 & \multirow{2}{*}{DST~\cite{liu2020dynamic}} & 80.4 & \multirow{2}{*}{DS} & 74.0 $\pm$0.41 & 76.8 / -2.8 & 67.1 & DS & 5.0 & 84.9 \\
 &  & 90.1 &  & 72.8 $\pm$0.27 & 76.8 / -4.0 & 75.8 & DS & 10.0 & 91.3 \\ \cline{2-10} 
 & \multirow{2}{*}{SWAT-U} & \textbf{80.0} & \textbf{80.0} & \textbf{75.2$\pm$0.06} & 76.8 / \textbf{-1.6} & \textbf{76.1} & \textbf{39.2} & \textbf{5.0} & 77.7 \\
 &  & 90.0 & 90.0 & 72.1$\pm$0.03 & 76.8 / -4.7 & 85.6 & 44.0 & 10.0 & 87.4 \\ \hline

 & \multirow{2}{*}{SWAT-ERK} & \textbf{80.0} & \textbf{52.0} & \textbf{76.0$\pm$0.16} & 76.8 / \textbf{-0.8} & \textbf{60.0} & \textbf{25.5} & \textbf{5.0} & 58.9 \\
 &  & 90.0 & 64.0 & 73.8$\pm$0.23 & 76.8 / -3.0 & 79.0 & 31.4 & 10.0 & 77.8 \\ \hline

 & \multirow{2}{*}{SWAT-M} & \textbf{80.0} & \textbf{49.0} & \textbf{74.6$\pm$0.11} & 76.8 / \textbf{-2.2} & \textbf{45.9} & \textbf{23.7} & \textbf{5.0} & 45.0 \\
 &  & 90.0 & 57.0 & 74.0$\pm$0.18 & 76.8 / -2.8 & 65.4 & 27.2 & 10.0 & 64.8 \\ \hline

\multirow{2}{*}{WRN-50-2} & \multirow{2}{*}{SWAT-U} & 80.0 & 80.0 & 76.4$\pm$0.10 & 78.5 / -2.1 & 78.6 & 39.1 & 5.0 & 79.2 \\
 &  & 90.0 & 90.0 & 74.7$\pm$0.27 & 78.5 / -3.8 & 88.4 & 43.9 & 10.0 & 89.0 \\ \hline
\end{tabular}%
}
\end{table}

\autoref{table:comparisioncifar10unstructured} compares SWAT-U, SWAT-ERK and SWAT-M with unstructured sparsity 
versus published results for DST~\cite{liu2020dynamic} and SNFS~\cite{dettmers2019sparse} for VGG-16-D, WRN-16-8, and DenseNet-121 on CIFAR-10.
Under the heading ``Training Sparsity'' we plot the average sparsity for weights (W) and activations (Act).
For SWAT-ERK and SWAT-M per layer sparsity for weights and activations are equal for a given layer but their
averages for the entire network differ because these are computed by weighting by the number of weights or activations per layer.
Comparing SWAT against SNFS and DST for $\text{VGG-16}$ and $\text{WRN-16-8}$ the data shows
that $\text{SWAT-M}$ has better accuracy versus SNFS and DST on both networks.
While $\text{SWAT-M}$ reduces training FLOPs by 33\% and 22\% versus SNFS and DST, respectively, 
$\text{SWAT-M}$ requires more training FLOPs versus DST on $\text{VGG-16}$. 
SWAT-U has better accuracy versus SNFS and DST on $\text{WRN-16-8}$ and 
$\text{SWAT-U}$ obtains $2.53\times$ and $1.96\times$ harmonic mean reduction in remaining training FLOPS versus SNFS and DST, respectively.  
\textcolor{black}{It is important to note that the reduction in FLOPs for SWAT-ERK is competitive with SWAT-U. In general, uniform 
will require fewer training and inference computations verses
ERK when the input resolution is high since ERK generally applies lower sparsity at the initial layer resulting in significant initial overhead. 
However, in \autoref{table:comparisioncifar10unstructured}, for all these networks the initial layer will have less computation due to the small input resolution of the CIFAR-10 dataset, and computationally expensive layers are 
allotted higher sparsity in SWAT-ERK.}

The data under ``Act $\downarrow$'' report the fraction of activation elements masked to zero, which can be
exploited by hardware compression proposed by NVIDIA~\cite{rhu2018compressing} to reduce transfer time between GPU and CPU.

\begin{wrapfigure}{r}{4.5cm}
\vspace{-0.3in}
\includegraphics[width=4.5cm]{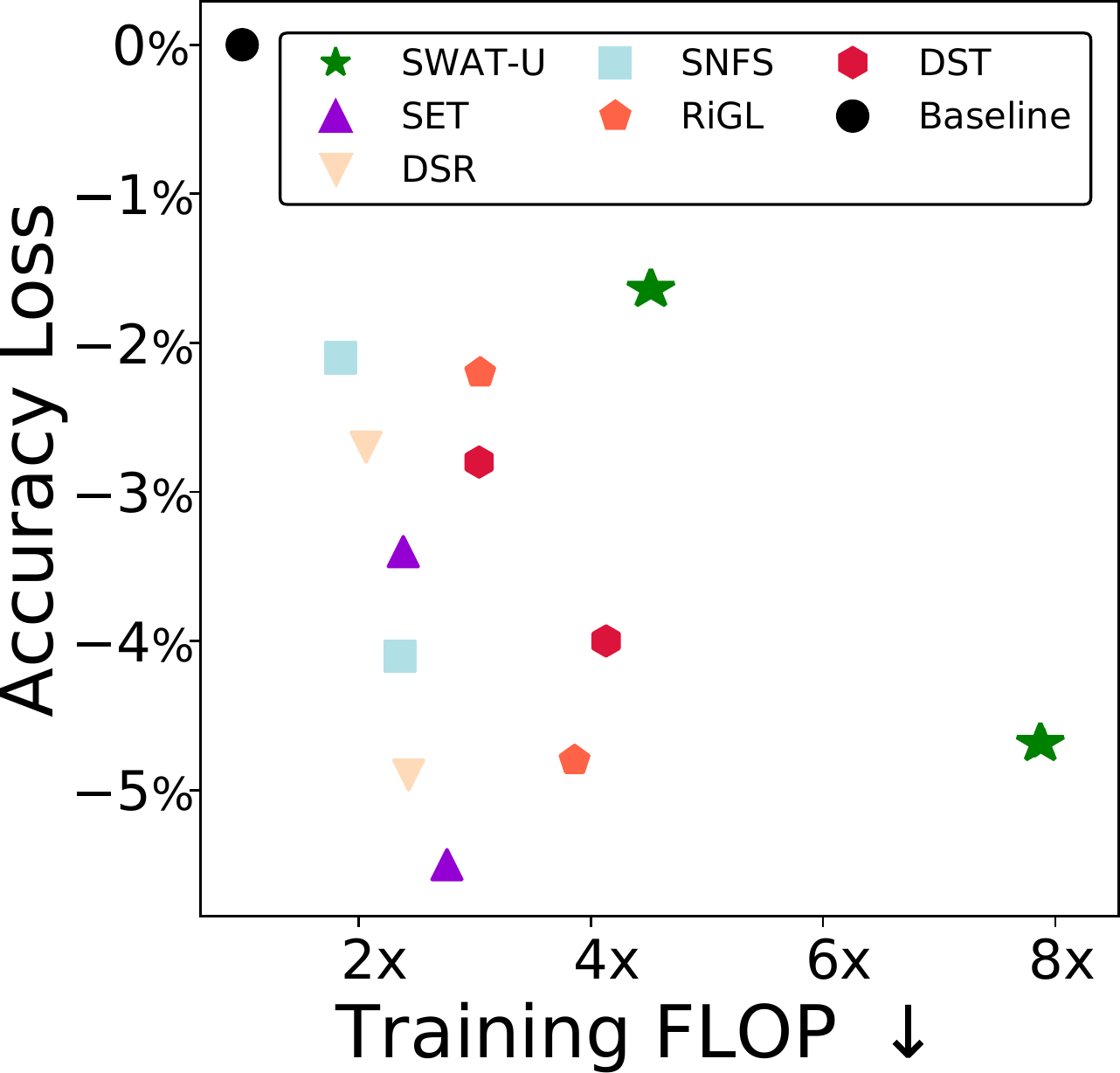}
\vspace{-0.2in}
\caption{Accuracy decrease versus reduction in training FLOPs}
\label{fig:scatter-plot}
\end{wrapfigure}

\paragraph{ImageNet:}
\autoref{table:comparisionimagenetunstructured} compares SWAT-U with unstructured sparsity 
against six recently proposed sparse learning algorithms 
at target weight sparsities of 80\% and 90\% on ImageNet.
Data for all sparse learning algorithms except RigL 
were obtained by running their code using the hyperparameters in Section~\ref{id}. 
Results for RigL are quoted from \citet{evci2019rigging}. 
SET and DSR do not sparsify downsampling layers leading to increased parameter count. 
At 80\% sparsity, SWAT-U attains the highest validation accuracy while reducing training FLOPs.
\textcolor{black}{DST trains ResNet-50 on ImageNet dataset with less computation~(``Training FLOPs'') than RigL even though DST is a dense to sparse training algorithm 
while RigL is a sparse to sparse training algorithm. This occurs for two reasons: (1) DST quickly reaches a relatively sparse 
topology after a few initial epochs; in our experiment, DST discards more than 70\% of network parameter within 5 epochs; 
(2) The sparsity distribution across layers is the crucial factor deciding reduction in FLOPs since allocating higher sparsity 
to the computationally expensive layer alleviates the initial overhead during entire training. Therefore, the overall benefit of DST is 
dependent on network architecture,  sparsity distribution, and the parameter discard rate.}
\autoref{fig:scatter-plot} plots the reduction in validation accuracy versus the reduction in training FLOPs 
relative to baseline dense training at both 80\% and 90\% sparsity using the same data 
as \autoref{table:comparisionimagenetunstructured}.  This figure shows
SWAT-U provides the best tradeoff between validation and reduction in training FLOPs.

\subsection{Structured SWAT}
\paragraph{CIFAR10:}
\autoref{table:comparision_cifar10_structured} provides results for SWAT-U with channel pruning  
on the CIFAR-10 dataset. 
At 70\% sparsity SWAT-U with channel pruning improves validation accuracy on both ResNet-18 and DenseNet-121 while reducing training FLOPs by $3.3\times$.

\begin{table}[]
	\caption{Structured SWAT on the CIFAR-10 dataset. \textbf{W:} Weight, \textbf{Act:} Activation, \textbf{CP:} Channel Pruned, \textbf{BA:} Baseline Accuracy, \textbf{AC:} Accuracy Change, \textbf{MC:} Model Compression.}
\label{table:comparision_cifar10_structured}
\resizebox{\textwidth}{!}{%
\begin{tabular}{cccccccccc}
\hline
\multirow{3}{*}{Network} & \multirow{3}{*}{Methods} & \multicolumn{2}{c}{Training Sparsity} & \multicolumn{2}{c}{Top-1} & \multicolumn{2}{c}{\bf Training} & \multicolumn{2}{c}{Inference} \\ \cline{3-10}
 &  & \multirow{2}{*}{\begin{tabular}[c]{@{}c@{}}W (\%) /\\ Act (\%)\end{tabular}} & \multirow{2}{*}{\begin{tabular}[c]{@{}c@{}}CP\\  (\%)\end{tabular}} & \multirow{2}{*}{\begin{tabular}[c]{@{}c@{}}Acc. $\pm$ SD ($\sigma$)\\  (\%)\end{tabular}} & \multirow{2}{*}{\begin{tabular}[c]{@{}c@{}}BA /\\ AC\end{tabular}} & \multirow{2}{*}{\begin{tabular}[c]{@{}c@{}}\bf FLOPs $\downarrow$\\  (\%)\end{tabular}} & \multirow{2}{*}{\begin{tabular}[c]{@{}c@{}}Act $\downarrow$\\  (\%)\end{tabular}} & \multirow{2}{*}{\begin{tabular}[c]{@{}c@{}}MC\\  ($\times$)\end{tabular}} & \multirow{2}{*}{\begin{tabular}[c]{@{}c@{}}FLOPs $\downarrow$\\ (\%)\end{tabular}} \\
 &  &  &  &  &  &  &  &  &  \\ \hline
\multirow{3}{*}{ResNet-18} & \multirow{3}{*}{SWAT-U} & 50.0/50.0 & 50.0 & 94.73$\pm$0.06 & 94.59 / +0.14 & 49.8 & 26.0 & 2.0 & 49.8 \\
 &  & 60.0/60.0 & 60.0 & 94.68$\pm$0.03 & 94.59 / +0.09 & 59.8 & 31.2 & 2.5 & 59.8 \\
 &  & 70.0/70.0 & 70.0 & 94.65$\pm$0.19 & 94.59 / +0.06 & 69.8 & 36.4 & 3.3 & 69.8 \\ 
	\hline
\multirow{3}{*}{DenseNet-121} & \multirow{3}{*}{SWAT-U} & 50.0/50.0 & 50.0 & 95.04$\pm$0.26 & 94.51 / +0.53 & 49.9 & 25.0 & 2.0 & 49.9 \\
 &  & 60.0/60.0 & 60.0 & 94.82$\pm$0.11 & 94.51 / +0.31 & 59.9 & 30.0 & 2.5 & 59.9 \\
 &  & 70.0/70.0 & 70.0 & 94.81$\pm$0.20 & 94.51 / +0.30 & 69.9 & 35.0 & 3.3 & 69.9 \\ 
	\hline
\end{tabular}%
}
\end{table}

\paragraph{ImageNet:}
\autoref{table:comparision_imagenet_structured} compares SWAT-U with channel pruning against 
four recent structured pruning 
algorithms. 
At 70\% sparsity SWAT-U with channel pruning reduces training FLOPs by $3.19\times$ by  
pruning 70\% of the channels on ResNet50 while incurring only $1.2\%$ loss in validation accuracy. 
SWAT-U with structured sparsity shows better accuracy but shows larger drops versus our baseline (trained with label smoothing) 
in some cases.  The works we compare with start with a densely trained network, prune channels, then fine-tune and so
increase training time contrary to our objective.
\begin{table}[]
	\caption{Structured SWAT on the ImageNet dataset.  \textbf{W:} Weight, \textbf{Act:} Activation, \textbf{CP:} Channel Pruned, \textbf{BA:} Baseline Accuracy, \textbf{AC:} Accuracy Change, \textbf{MC:} Model Compression.}
\label{table:comparision_imagenet_structured}
\resizebox{\textwidth}{!}{%
\begin{tabular}{cccccccccc}
\hline
\multirow{3}{*}{Network} & \multirow{3}{*}{Methods} & \multicolumn{3}{c}{Training} & \multicolumn{2}{c}{Pruning} & \multicolumn{2}{c}{Top-1} & Inference \\ \cline{3-10} 
 &  & \multirow{2}{*}{\begin{tabular}[c]{@{}c@{}}W (\%) /\\ Act (\%)\end{tabular}} & \multirow{2}{*}{\begin{tabular}[c]{@{}c@{}}FLOPs $\downarrow$\\  (\%)\end{tabular}} & \multirow{2}{*}{\begin{tabular}[c]{@{}c@{}}Act $\downarrow$\\  (\%)\end{tabular}} & \multirow{2}{*}{\begin{tabular}[c]{@{}c@{}}CP\\  (\%)\end{tabular}} & \multirow{2}{*}{\begin{tabular}[c]{@{}c@{}}Fine-Tune\\  (epoch)\end{tabular}} & \multirow{2}{*}{\begin{tabular}[c]{@{}c@{}}Acc\\  (\%)\end{tabular}} & \multirow{2}{*}{\begin{tabular}[c]{@{}c@{}}BA / \\ AC\end{tabular}} & \multirow{2}{*}{\begin{tabular}[c]{@{}c@{}}FLOPs $\downarrow$\\  (\%)\end{tabular}} \\
 &  &  &  &  &  &  &  &  &  \\ \hline
\multirow{7}{*}{ResNet-50} & DCP~\cite{zhuang2018discrimination} & \multicolumn{3}{c}{Offline Pruning} & - & 60 & 74.95 & 76.01 / -1.06 & 55.0 \\
 & CCP~\cite{peng2019collaborative} & \multicolumn{3}{c}{Offline Pruning} & 35 & 100 & 75.50 & 76.15 / -0.65 & 48.8 \\
 & AOFP~\cite{ding2019approximated} & \multicolumn{3}{c}{Offline Pruning} & - & Yes (-) & 75.11 & 75.34 / -0.23 & 56.73 \\
 & Soft-Pruning~\cite{he2018soft} & - & - & - & 30 & No & 74.61 & 76.15 / -1.54 & 41.8 \\
 & \multirow{3}{*}{\begin{tabular}[c]{@{}c@{}}SWAT-U\\ (Structured)\end{tabular}} & 50.0/50.0 & 47.6 & 24.5 & 50 & No & 76.51$\pm$0.30 & 76.80 / -0.29 & 48.6 \\
 &  & 60.0/60.0 & 57.1 & 29.5 & 60 & No & 76.35$\pm$0.06 & 76.80 / -0.45 & 58.3 \\
 &  & 70.0/70.0 & 66.6 & 34.3 & 70 & No & 75.67$\pm$0.06 & 76.80 / -1.13 & 68.0 \\ \hline
\multirow{3}{*}{WRN-50-2} & \multirow{3}{*}{\begin{tabular}[c]{@{}c@{}}SWAT-U\\ (Structured)\end{tabular}} & 50.0/50.0 & 49.1 & 24.5 & 50 & No & 78.08$\pm$0.20 & 78.50 / -0.42 & 49.5 \\
 &  & 60.0/60.0 & 58.9 & 29.4 & 60 & No & 77.55$\pm$0.07 & 78.50 / -0.95 & 59.4 \\
 &  & 70.0/70.0 & 68.7 & 34.2 & 70 & No & 77.19$\pm$0.11 & 78.50 / -1.31 & 69.3 \\ \hline
\end{tabular}%
}
\end{table}

\subsection{Sparse Accelerator Speedup} 
\textcolor{black}{We believe SWAT is well suited for emerging sparse machine learning accelerator hardware designs like NVIDIA's recently announced Ampere (A100) GPU. 
To estimate the speedup on a sparse accelerator, we have used an architecture simulator developed for a recent sparse accelerator hardware proposal ~\cite{delmas2019bit}. 
The simulator counts only the effectual computation (non-zero computation)  and exploits the sparsity present in the computation. 
The architecture has a 2D array of processing units where each processing unit has an array of multipliers and dedicated weight and activation and accumulation buffers. 
It counts the cycles taken to spatially map and schedules the computation present in each layer of the network.
The memory hierarchy is similar to the DaDianNo architecture~\cite{chen2014dadiannao}. The activation and weight buffer can hold one entire layer at a time and hide the latency of memory transfer. 
The memory throughput is high enough to satisfy the computation throughput.} 

\textcolor{black}{The simulator only implements the forward pass, and therefore the simulator does not simulate the backward pass. However, the backward pass convolution is a 
transposed convolution, which can be translated into a standard convolution by rotating the input tensor~\cite{dumoulin2016guide}. 
So we estimated the backward pass speedup by transforming transpose convolution into a standard convolution and using the inference simulator. 
Note, here the transformation overhead was not considered in the speedup. However, the overhead would be small in the actual hardware 
since this transformation operation is a simple rotation operation, i.e., rotating the tensor along some axis and, therefore, we assume it could be accelerated on hardware. 
\begin{table}[]
\caption{Speedup due to SWAT on ML Accelerator}
\label{tab:accelerator}
\centering
\begin{tabular}{@{}ccc@{}}
\toprule
Top-$K$ Sparsity & Forward Pass Speed Up & Backward Pass Speed Up \\ \midrule
0\%  & 1$\times$ & 1$\times$ \\
80\%  & 3.3$\times$ & 3.4$\times$ \\
90\%  & 5.6$\times$ & 6.3$\times$ \\ \bottomrule
\end{tabular}
\end{table}
\autoref{tab:accelerator} reports forward and backward pass training speedup (in simulated clock cycles) for SWAT-U with unstructured pruning on ResNet-50 with ImageNet.
From \autoref{table:comparisionimagenetunstructured} we see that at 80\% sparsity SWAT-U incurs 1.63\% accuracy loss but the data in \autoref{tab:accelerator} suggests it
may improve training time by $3.3\times$ on emerging platforms supporting supporting hardware acceleration of sparse computations.
}

\subsection{Effect of updates to non-active weights} 
\label{sec:ablation}
\begin{figure}[H]
\centering
\begin{minipage}{.46\linewidth}
\centering
  \vspace{-0.1in}
\includegraphics[scale=0.23]{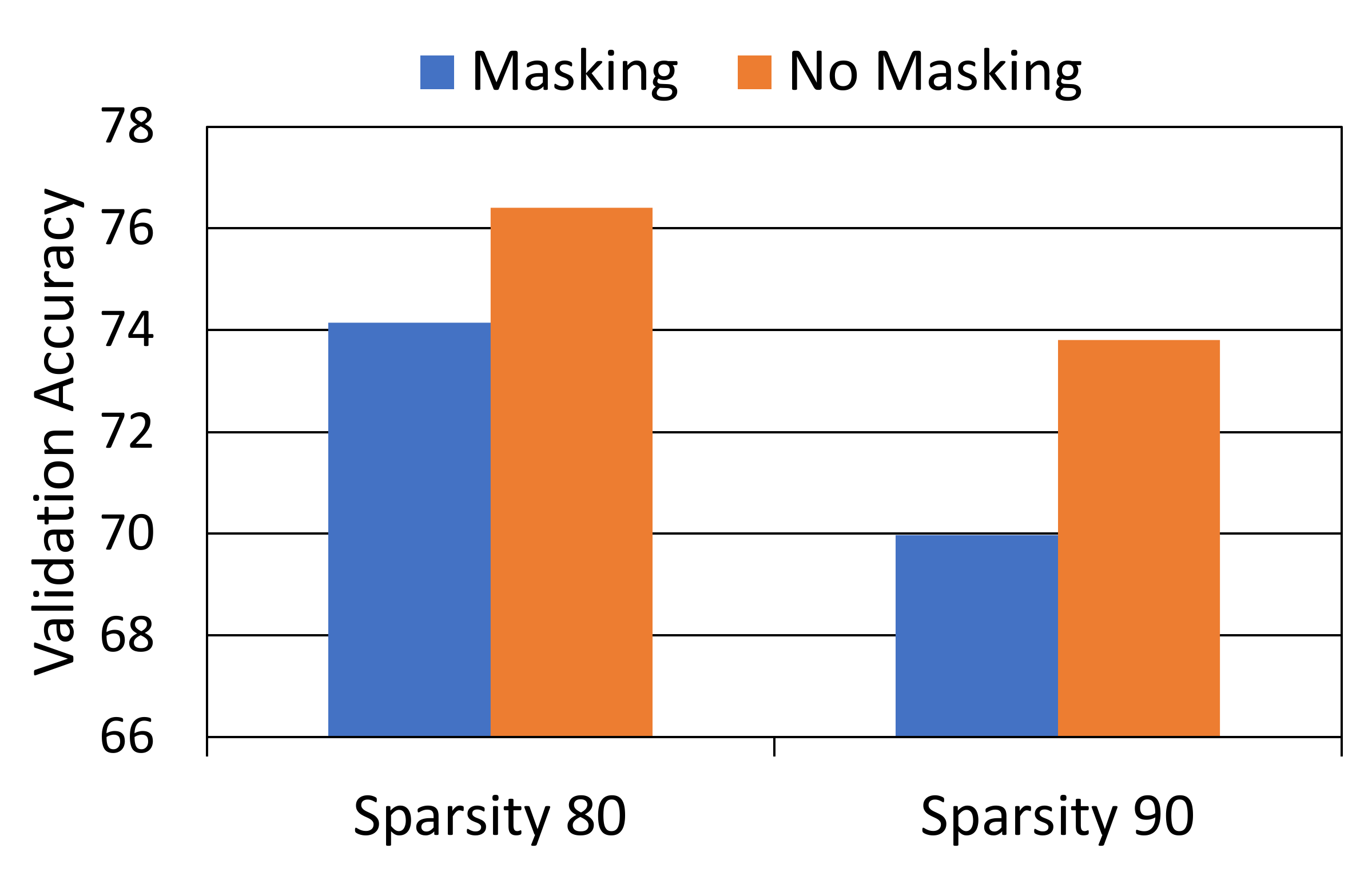}
  \vspace{-0.1in}
\caption{Effect of masking of gradient update for non-active weights.} 
    \label{fig-masking}	
\end{minipage}
\hfill
\noindent\begin{minipage}{.48\linewidth}
\centering
\includegraphics[scale=0.23]{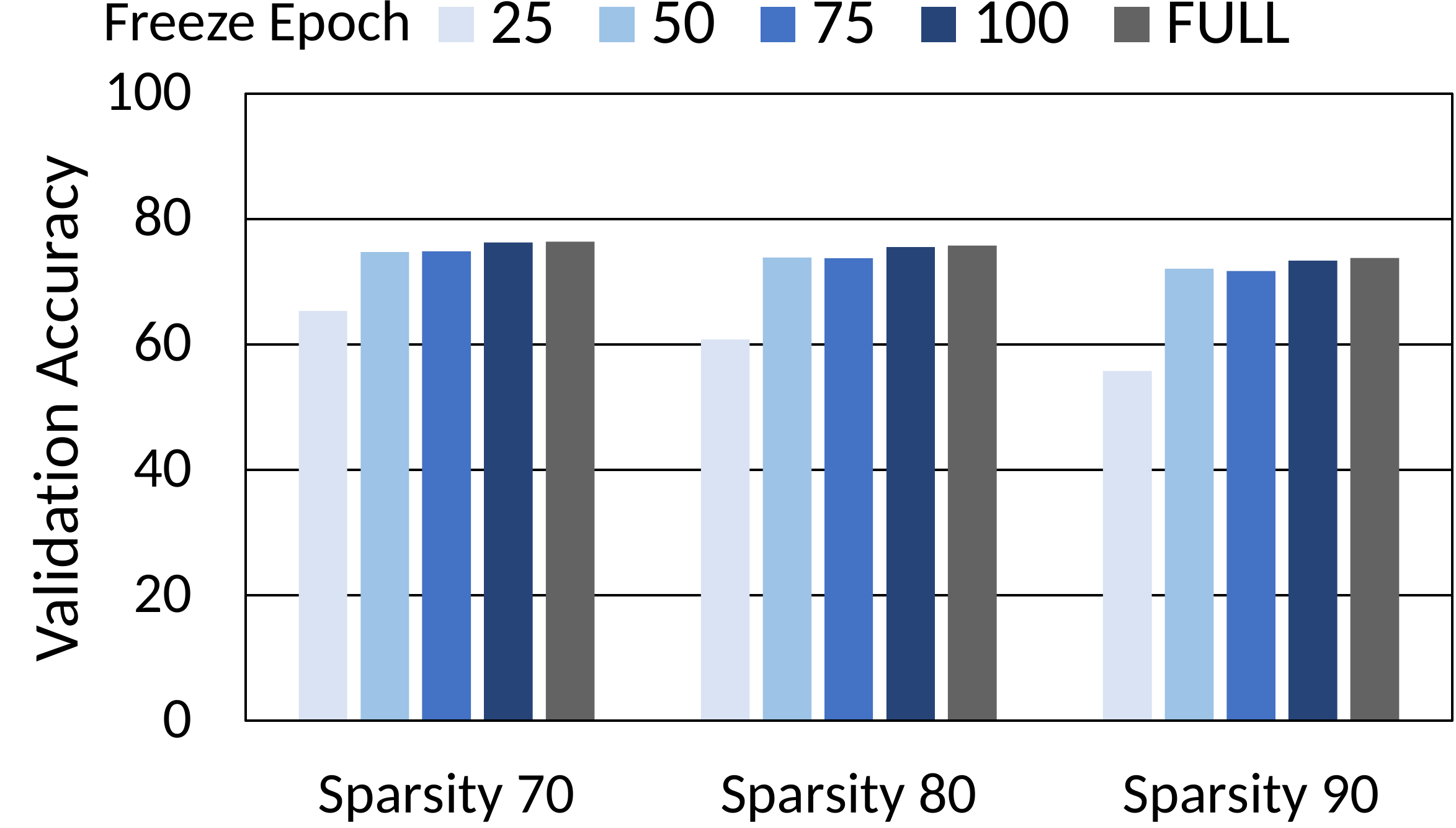}
\caption{Effect of stopping dynamic topology exploration early.} 
\label{fig-freezeepoch}
\end{minipage}%
\end{figure}

SWAT updates non-active weights in the backward pass using dense gradients and one might reasonable ask
whether it would be possible to further decrease training time by masking these gradients.
\autoref{fig-masking} measures the effect of this masking on validation accuracy 
while training ResNet-18 on CIFAR-100.  The data suggest it is important to update non-active weights.
Since non-active weights are updated they may become active changing the topology.  
\autoref{fig-freezeepoch} shows the effect of freezing the topology after differing number of epochs during training of ResNet18 on CIFAR100 dataset. 
Freezing topology exploration early is harmful to convergence and results in reduced final validation accuracy.

\section{Conclusion}
In this work, we propose SWAT, a novel efficient training algorithm that sparsifies both the forward and the backward passes with negligible impact on convergence. SWAT is the first sparse learning algorithm we are aware of that can train both structured and unstructured sparse networks. 
SWAT achieves validation accuracy comparable to other pruning and sparse learning algorithms while demonstrating potential for significant reduction in 
training time on emerging platforms supporting hardware acceleration of sparse computations. 

\section{Broader Impact}
This work has the following potential positive impact in society: 1) It is an entirely sparse training algorithm for emerging sparse hardware accelerators. 2) It makes training efficient and faster. Thus, decreasing the model training cost. 3) It can reduce the overall carbon footprint of model training, which is a huge problem. \citet{strubell2019energy} shows that the carbon footprint of training models can release  $5\times$ the carbon emission of a car during its lifetime.  4)  It can enable us to train even bigger models, thus allowing us to achieve new state-of-the-art accuracies.

 At the same time, this work may have some negative consequences because SWAT can enable us to develop better AI and better AI technologies may have negative societal implications such as strict surveillance, privacy concerns, and job loss.

\section{Acknowledgements}
We thank Francois Demoullin, Negar Goli, Dave Evans, Deval Shah, Yuan Hsi Chou, and the anonymous reviewers for their valuable comments on this work. 
This research has been funded in part by the Computing Hardware for Emerging Intelligent Sensory Applications (COHESA) project. 
COHESA is financed under the National Sciences and Engineering Research Council of Canada (NSERC) Strategic Networks grant number NETGP485577-15.

\bibliography{neurips_2020}
\bibliographystyle{plainnat}

\begin{appendices}
\section{Appendix}
\subsection{Detailed description of the SWAT algorithm}\label{Algorithm}

The SWAT algorithm is summarized in Algorithm~\ref{algorithm}. The shaded
region represents the sparse computation step which could be exploited on
sparse machine learning accelerators. The SWAT algorithm consists of three
parts: Forward Computation, Backward Computation, and the Parameter Update. 

\begin{algorithm}[]
\SetAlgoLined
\caption{SWAT Algorithm}\label{algorithm} 
\label{Algorithm}
  \KwData{Training iteration $t$ and Top-$K$ sampling period $P$.  
  Network with $L$ layers and previous weight parameters $\textbf{w}^t$.  
  Sparsity distribution algorithm $D$.
  Mini-batch of inputs ($\textbf{a}_0$) and corresponding targets ($\textbf{a}^*$). 
  Learning rate $\eta$. Gradient descent optimization algorithm $Optimizer$.}
\KwResult{Updated weight parameters $\textbf{w}^{t+1}$.} 
\KwComment{}
\KwComment{Stage 1. Forward Computation\;}
\For{$l$ = 1 to $L$} 
{
  \label{s1:begin}

  \eIf{$l$ is a convolutional or fully-connected layer}{ \label{s1:ifcond}

  \If{ $t$ mod $P$ = 0 } {
  $S_l \Leftarrow \texttt{getLayerSparsity(}l,D\texttt{)}$\; \label{s1:getLayerSparsity}

  $t^w_l \Leftarrow \texttt{getThreshold(}\textbf{w}_l\texttt{,}S_l\texttt{,}t\texttt{)}$\; \label{s1:getThresholdW}

  $t^a_l \Leftarrow \texttt{getThreshold(}\textbf{a}_{l-1}\texttt{,}S_l\texttt{,}t\texttt{)}$\; \label{s1:getThresholdA}
  }

    $\textbf{W}_l^{t} \Leftarrow  f_{TOPK}(\textbf{w}_l^t,t^w_l)$\; \label{s1:topk:w}

\tikzmk{A}
  $\textbf{a}_l^t \Leftarrow \texttt{forward}(\textbf{W}_l^{t}, \textbf{a}_{l-1}^t)$\; \label{s1:forward}
\tikzmk{B}
\boxit{blue}
    
  $\textbf{a}_{l-1}^{t} \Leftarrow  f_{TOPK}(\textbf{a}_{l-1}^t,t^a_l)$\; \label{s1:topk:a}

$\textbf{save\_for\_backward}_l\Leftarrow \textbf{W}_l^{t}, \textbf{a}_{l-1}$\; \label{s1:save}
}
{
  $\textbf{a}_l \Leftarrow \texttt{forward}(\textbf{w}_l^{t}, \textbf{a}_{l-1})$\; \label{s1:normal_fwd}

  $\textbf{save\_for\_backward}_l\Leftarrow \textbf{w}_l^{t}, \textbf{a}_{l-1}$\; \label{s1:save_dense}
}
} \label{s1:end}
\KwComment{}
\KwComment{Stage 2. Backward Computation\;}
Compute the gradient of the output layer ${\bigtriangledown}_{\textbf{a}_L}=\frac{\partial loss(\textbf{a}_L, \textbf{a}^*)}{\partial \textbf{a}_L}$\; \label{s2:loss} 
\KwComment{}
\For{$l$ = L to 1}
{
  \label{s2:begin}
  $\textbf{W}_l^{t}, \textbf{a}_{l-1}\Leftarrow \textbf{save\_for\_backward}_l$\; \label{s2:fetch}

\eIf{$l$ is a convolutional or fully-connected layer}
{
	\tikzmk{A}
	${\bigtriangledown}_{\textbf{a}_{l-1}} \Leftarrow \texttt{backward\_input}({\bigtriangledown}_{\textbf{a}_{l}},\textbf{W}_l^{t})$\;
	\tikzmk{B}
	\boxit{blue}
	
	\tikzmk{A}
	${\bigtriangledown}_{\textbf{w}_{l-1}} \Leftarrow \texttt{backward\_weight}({\bigtriangledown}_{\textbf{a}_{l}},\textbf{a}_{l-1}^{t})$\;
	\tikzmk{B}
	\boxit{blue}
}
{
	${\bigtriangledown}_{\textbf{a}_{l-1}} \Leftarrow \texttt{backward\_input}({\bigtriangledown}_{\textbf{a}_{l}},\textbf{W}_l^{t})$\;
	
	${\bigtriangledown}_{\textbf{w}_{l-1}} \Leftarrow \texttt{backward\_weight}({\bigtriangledown}_{\textbf{a}_{l}},\textbf{a}_{l-1}^{t})$\;
		
}
	
} \label{s2:end}
\KwComment{}
\KwComment{Stage 3. Parameter Update\;}
\For{$l$ = 1 to L}
{ \label{s3:begin}
$\textbf{w}_l^{t+1}\Leftarrow Optimizer(\textbf{w}_l^t, {\bigtriangledown}_{\textbf{w}_{l}},\eta)$\;
} \label{s3:end}
\end{algorithm}

The forward computation~(Line~\ref{s1:begin} to \ref{s1:end}) for each layer proceeds as follows:  
First, we check if the current layer is a convolutional or fully-connected layer (Line~\ref{s1:ifcond}).
If neither, perform the regular (non-SWAT) forward pass computation (Line~\ref{s1:normal_fwd}) and save dense weights and activations (Line~\ref{s1:save_dense}).
Otherwise, if the training iteration $t$ is a multiple of the Top-$K$ sampling period $P$ then
we obtain the target sparsity {\bf S}$_l$ of layer~$l$ based upon distribution algorithm $D$ (Line~\ref{s1:getLayerSparsity})
where $D$ is one of the techniques described in Section~3.3.2. 
Then, we compute threshold $t^w$ for weight sparsity~(Line~\ref{s1:getThresholdW}) used in the forward pass
and threshold $t^a$ for use in sparsifying activations prior to saving to memory for use in the the backward pass~(Line~\ref{s1:getThresholdA}).
Prior to performing the forward computation (Line~\ref{s1:forward}) we compute the active weights $\textbf{W}_l^{t}$ by 
applying the sparsifying function, $f_{TOPK}$ using threshold $t^w$. For input tensor \textbf{x}$_i$ 
For fine-grained sparsity $f_{TOPK}($\textbf{x},$t)$ maps input elements 
$x_i \in \textbf{x}$ to output elements $y_i$ according to:
\begin{equation*}
  y_i = 
\begin{cases}
  0, & |x_i| \leq t \\
  x_i, & \text{otherwise}
\end{cases}
\end{equation*}
For coarse-grained sparsity (Section~3.3.1), which we apply only to weights, $f_{TOPK}($\textbf{x},$t)$ maps input elements 
$x_i \in \textbf{x}$ to output elements $y_i$ where $i$ is an element of channel $C$ 
according to:
\begin{equation*}
  y_i = 
\begin{cases}
  0, & \sum_{i \in C} |x_i| \leq t \\
  x_i, & \text{otherwise}
\end{cases}
\end{equation*}
Next, we perform sparse forward computations, \texttt{forward}, corresponding to Equation~1 in the paper to generate output activation~(Line~\ref{s1:forward}).
Next, we apply fine-grained Top-$K$ sparsification to the input activations~(Line~\ref{s1:topk:a}).
Save sparse active weight parameters $\textbf{W}_l^{t}$ and input activations $\textbf{a}_{l-1}$ for the backward pass~(Line~\ref{s1:save}).

After the forward pass the loss function is applied and the back propagated error-gradient for the output layer is computed (Line~\ref{s2:loss}). 
Then, the backward pass computation~(Line~\ref{s2:begin} to \ref{s2:end}) proceeds as follows:
First, we load the saved parameters and input activations of the current layer~(Line~\ref{s2:fetch}). 
Next, we perform the backward pass to generate the input activation gradients and weight gradients using \texttt{backward\_input} and \texttt{backward\_weight}
functions, which correspond to Equations~2 and 3 in the paper, respectively. 
As sparse input activations and parameters are saved in the forward pass 
the computation is sparse.

After the backward pass of the current mini-batch, the optimizer uses the computed weight gradients to update the parameters~(Line~\ref{s3:begin} to \ref{s3:end}).

\subsection{CIFAR-100}
\subsubsection{Unstructured SWAT}
\autoref{cifar100unstructuredswat} compares SWAT-U, SWAT-ERK and SWAT-M with unstructured sparsity for VGG-16, WRN-16-8 and DenseNet-121 architecture on CIFAR-100 dataset. 
The training procedure is the same as outlined in Section~4.1 in the paper.  Hyperparameters are listed in Appendix \autoref{Hyperparameters}.
\begin{table}[h!]
\caption{Unstructured SWAT on CIFAR-100 dataset.}
\label{cifar100unstructuredswat}
\resizebox{\textwidth}{!}{%
\begin{tabular}{cccccc}
\hline
\multirow{3}{*}{Network} & \multirow{3}{*}{Methods} & \multicolumn{2}{c}{Training Sparsity} & \multicolumn{2}{c}{Top-1} \\ \cline{3-6} 
 &  & \multirow{2}{*}{Weight (\%)} & \multirow{2}{*}{Activation (\%)} & \multirow{2}{*}{\begin{tabular}[c]{@{}c@{}}Accuracy\\  (\%)\end{tabular}} & \multirow{2}{*}{Accuracy Change} \\
 &  &  &  &  &  \\ \hline
\multirow{3}{*}{VGG-16} & SWAT-U & 90.0 & 90.0 & 69.8 & -2.3 \\
 & SWAT-ERK & 90.0 & 69.6 & 71.8 & -0.3 \\
 & SWAT-M & 90.0 & 59.9 & 72.2 & +0.1 \\ \cline{2-6} 
\multirow{3}{*}{WRN-16-8} & SWAT-U & 90.0 & 90.0 & 77.6 & -1.7 \\
 & SWAT-ERK & 90.0 & 77.6 & 78.5 & -0.8 \\
 & SWAT-M & 90.0 & 73.3 & 77.9 & -1.4 \\ \cline{2-6} 
\multirow{3}{*}{DENSENET-121} & SWAT-U & 90.0 & 90.0 & 77.2 & -0.4 \\
 & SWAT-ERK & 90.0 & 90.0 & 76.5 & -1.1 \\
 & SWAT-M & 90.0 & 84.2 & 75.5 & -2.1 \\ \cline{2-6} 
\end{tabular}%
}
\end{table}

\subsubsection{Structured SWAT}
\autoref{cifar100structuredswat} compares SWAT-U  with unstructured sparsity for ResNet-18 and DenseNet-121 architecture on CIFAR-100 dataset. 
The training procedure is the same as outlined in Section~4.1 in the paper.  Hyperparameters are listed in Appendix \autoref{Hyperparameters}.

\begin{table}[h!]
\caption{Structured SWAT on CIFAR-100 dataset.}
\label{cifar100structuredswat}
\resizebox{\textwidth}{!}{%
\begin{tabular}{ccccccc}
\hline
\multirow{3}{*}{Network} & \multirow{3}{*}{Methods} & \multicolumn{3}{c}{Training Sparsity} & \multicolumn{2}{c}{Top-1} \\ \cline{3-7} 
 &  & \multirow{2}{*}{Weight (\%)} & \multirow{2}{*}{Activation (\%)} & \multirow{2}{*}{\begin{tabular}[c]{@{}c@{}}Channel \\ Pruned (\%)\end{tabular}} & \multirow{2}{*}{\begin{tabular}[c]{@{}c@{}}Accuracy\\  (\%)\end{tabular}} & \multirow{2}{*}{Accuracy Change} \\
 &  &  &  &  &  &  \\ \hline
\multirow{3}{*}{RESNET-18} & \multirow{3}{*}{SWAT-U} & 50.0 & 50.0 & 50.0 & 76.4 & -0.4 \\
 &  & 60.0 & 60.0 & 60.0 & 76.2 & -0.6 \\
 &  & 70.0 & 70.0 & 70.0 & 75.6 & -1.2 \\ \cline{2-7} 
\multirow{3}{*}{DENSENET-121} & \multirow{3}{*}{SWAT-U} & 50.0 & 50.0 & 50.0 & 78.7 & +0.9 \\
 &  & 60.0 & 60.0 & 60.0 & 78.5 & +0.4 \\
 &  & 70.0 & 70.0 & 70.0 & 78.1 & +0.3 \\ \hline
\end{tabular}%
}
\end{table}
	
\newpage
\subsection{FLOP Calculation}
Consider a convolution layer with input tensor $X\in \mathbb{R}^{N\times C\times X \times Y}$ and weight tensor 
$W\in \mathbb{R}^{F\times C\times R \times S}$ to produce output tensor $O\in \mathbb{R}^{N\times F\times H \times W}$. Input tensor has $N$ samples; each sample has $C$ input channels of dimension $X\times Y$. Weight tensor has $F$ filters and each filters has $C$ channels of dimension $R\times S$. Output tensor has $N$ output samples and each sample has $F$ output channels of dimension $H\times W$.

During the forward pass, input tensor is convolved with weight tensor to produce output tensor. In contrast, in the backward pass, the error-gradient of output tensor is deconvolved with input and weight tensor to produce weight gradient and input gradient respectively.
The forward pass FLOP calculation assumes $s1$ sparsity in weight tensor. 
The effect of default sparsity in activation for forward pass computation is ignored since the default activation sparsity 
is present for both sparse learning and SWAT algorithms.
However, for the backward pass FLOP calculation, since for SWAT the activation is explicitly sparsified 
therefore the FLOP calculation for SWAT is done using $s1$ weight sparsity and $s2$ activation sparsity whereas for sparse learning algorithms, $s1$ weight sparsity and default activation sparsity is assumed. The default activation sparsity generally vary between $30-50\%$, for our calculation we assumed default activation sparsity of 50\%. 

All the sparse learning algorithms and SWAT require some extra FLOP for connectivity update and regrowth connection such as dropping low magnitude component and thresholding. We omit the FLOP needed for these operations in our training FLOP calculation. For dynamic sparse learning algorithms such as SNFS 
\cite{dettmers2019sparse}, DSR~\cite{mostafa2019parameter} and DST~\cite{liu2020dynamic}, the weight sparsity varies during iterations and therefore we computed the average weight sparsity for different layers during the entire training and used it for computing the training FLOP.

\subsubsection{Computation in Convolution Layer}

\textbf{Forward Pass}
\newline
Filter from the weight tensor is convolved with some sub-volume of input tensor of dimension $X^{sub}\in \mathbb{R}^{ C\times R \times S}$ to produce a single value in output tensor. Therefore, the total FLOP for any single value in output tensor is $C\times R\times S$ floating-point multiplication + $C\times R\times S - 1$ floating-point addition. This is approximately equal to $C\times R\times S$ floating-point MAC operations. Note, here 1 floating-point MAC operations = 1 floating-point multiplication + 1 floating-point addition. 
Thus, the total computation in the forward pass is equal to $ (C\times R\times S) \times N\times F\times H\times W$ MAC operations.

Now, lets assume weight tensor is sparse. The overall sparsity in the weight tensor is $s1$ and the sparsity per filter in the weight tensor is $s^{1},s^{2}$.....,$s^{F}$, for F filters in the layer. Note, $s1=\frac{\sum_{x=1}^{F} s^x}{F}$.
Theoretically only the non-zero weight will contribute to the FLOP. Therefore, the total FLOP for a single value in output tensor produce by the filter $x$ is $s^x\times C\times R\times S$ MAC operations. Thus, the total FLOP contribution for producing an output channel, of dimension $H\times W$, by the filter $x$ is $s^x\times (C\times R\times S) \times H\times W$. Therefore, the total FLOP for N input batches and F output channel is equal to $\sum_{x=1}^{F} (s^x\times (C\times R\times S) \times N \times H\times W ) = (\sum_{x=1}^{F} s^x) \times (C\times R\times S) \times N \times H\times W = s1\times F \times (C\times R\times S) \times N \times H\times W.$ Hence, theoretically the FLOP reduction is proportional to sparsity in weight tensor.

\begin{equation}
	\textbf{Forward Pass FLOP} = s1\times F \times (C\times R\times S) \times N \times H\times W.
\end{equation}
\textbf{Backward Pass}
\newline
For the backward pass, we back-propagate the error signal for computing the gradient of parameters. During the backward pass, each layer calculates 2 quantities: input gradient and weight gradient. 

For the input gradient computation, the output gradient is deconvolved with filters to produce the input gradient. 
It can be implemented by rotating the filter tensor from 
$W\in \mathbb{R}^{F\times C\times R \times S}$ to $W\in \mathbb{R}^{C\times F\times R \times S}$ and convolving with output gradient. 
In this computation, the convolution data is output gradient and convolution kernels are filters. 
Therefore, as described in the previous section, the total MAC in the input gradient is approximately proportional
to  $ (F\times R\times S) \times N \times C\times X\times Y$. Hence, the FLOP reduction with sparsity $s1$ in weight tensor will be approximately proportional to $s1\times C \times (F\times R\times S) \times N \times X\times Y$.

\begin{equation}
	\textbf{Input Gradient FLOP} = s1\times C \times (F\times R\times S) \times N \times X\times Y. 
\end{equation}
For the weight gradient computation, the input activation is deconvolved with the output gradient to produce the weight gradient. 
It can be implemented by rotating the output gradient tensor from $\bigtriangledown O\in \mathbb{R}^{N\times F\times H \times W}$ to $\bigtriangledown O\in \mathbb{R}^{F\times N\times H \times W}$ and 
input actvation from $X\in \mathbb{R}^{N\times C\times X \times Y}$ $X\in \mathbb{R}^{C\times N\times X \times Y}$. 
The rotated output gradient is convolved on input activation to produce weight gradient. In this computation, the convolution data is input activation and convolution kernel is output gradient. 
Therefore, the total MAC in the weight gradient is approximately proportional to  $ (N\times X\times Y) \times F \times C\times R\times S$. Hence, as described in the previous section, the FLOP reduction with sparsity $s2$ in input activation tensor will be approximately proportional to $s2\times (N\times X\times Y) \times F \times C\times R\times S$. 

\begin{equation}
	\textbf{Weight Gradient FLOP} = s2\times (N\times X\times Y) \times F \times C\times R\times S. 
\end{equation}

Thus, the computational expense of the backward pass is approximately twice that of forward pass.
\subsubsection{Computation in Linear Layer}\label{app:linear}
Consider a linear layer with input tensor $X\in \mathbb{R}^{N\times X}$ and weight tensor 
$W\in \mathbb{R}^{X\times Y}$ to produce output tensor $O\in \mathbb{R}^{N\times Y}$.
During the forward pass, input tensor is multiplied with weight tensor to produce output tensor. In contrast, in the backward pass, the error-gradient of output tensor is multiplied with input and weight tensor to produce error-gradient of weight and error-gradient input tensor.

\textbf{Forward Pass}
\newline
The total FLOP for the forward pass is $N\times Y\times X$ floating-point multiplication + $N\times Y\times (X - 1)$ floating-point addition. This is approximately equal to $N\times X\times Y$ floating-point MAC operations.

Now, lets assume weight tensor is sparse. The overall sparsity in the weight tensor is $s$ and the sparsity per column in the weight tensor is $s^{1},s^{2}$.....$s^{Y}$ for Y columns in the weight tensor. Note, $s=\frac{\sum_{y=1}^{Y} s^y}{Y}$.
Theoretically only the non-zero weight will contribute to the FLOP. Therefore, the total FLOP for a single value in the $y$ column of output tensor is $s^y\times X$ MAC operations. Thus, the total FLOP for all the element in the $y$ column of output tensor is $s^y\times X\times N$. Therefore, the FLOP in forward pass is equal to $\sum_{y=1}^{Y} s^y\times N\times Y=N\times X\times (\sum_{x=1}^{F} s^x) = s\times N\times X\times Y.$ Thus, theoretically the FLOP reduction will be proportional to sparsity in weight tensor. 

\textbf{Backward Pass}
\newline
The computational expense of the backward pass is twice that of forward pass and is reduced proportionally to the sparsity of weight and activation tensor.

\subsection{Top-K Overhead}
The Top-K operation can be efficiently implemented by finding the K-th largest element using an introselect algorithm and performing thresholding operation over the tensor. To estimate the overhead of finding the K-th largest element, we use numpy.partition function. The numpy.partition function uses the introselect algorithm and rearranges the array such that the element in the K-th position is in the position it would be in the sorted array. This overhead would be larger than the cost of finding the K-th largest element due to the extra rearrangement operations and thus serves as an upper bound on Top-K overhead.

We profile the numpy partition function using Vtune 2020.1.0.607630 on Intel(R) Core(TM) i9-7920X CPU @ 2.90GHz. \autoref{KthLargestElement} shows the total number of retired instruction while executing numpy.partition for different array size. We can observe that the overhead of the numpy.partition increases linearly with array size.
\begin{table}[h!]
\centering
\caption{Overhead of computing the K-th largest element using introselect algorithm}
\label{KthLargestElement}
\begin{tabular}{@{}ccc@{}}
\toprule
Array Size & Retired Instruction & Retired Instruction/Size \\ \midrule
1000 & 15215 & 15.2 \\
10000 & 217863 & 21.8 \\
100000 & 1381744 & 13.8 \\
1000000 & 12454369 & 12.6 \\ \bottomrule
\end{tabular}%
\end{table}

The Top-K function in Pytorch framework is not an optimized implementation for GPU. 
Therefore, to estimate the Top-K overhead on GPU, we used the efficient Radix based Top-K 
implementation by \citet{shanbhag2018efficient}. \autoref{TopKOverhead} shows the overall 
overhead of performing the Top-K operation for ResNet-50 on the ImageNet dataset with a 
batch-size of 32 on NVIDIA RTX-2080 Ti.

\begin{table}[h!]
\centering
\caption{Top-K Overhead}
\label{TopKOverhead}
\resizebox{\textwidth}{!}{%
\begin{tabular}{cccc}
\hline
Top-K Weight & Top-K Activation & Forward + Backward Pass & Total Top-K Overhead \\ \hline
18 ms & 452 ms & 147 ms & 470 ms \\ \hline
\end{tabular}%
}
\end{table}

SWAT-U reduces the training FLOP per iteration by 76.1\% and 85.6\% at 80\% and 90\% sparsity respectively. Therefore, theoretical training speed up (assuming hardware can directly translate FLOPs reduction into reduced execution time) for the SWAT-U algorithm with a Top-K period of 1000 iteration, at 80\%, and 90\% sparsity would be $4.13\times$ and $6.79\times$ respectively.  These numbers are found as follows:
\begin{equation}
	\text{Speed Up at 80\% sparsity} = \frac{1000 \times 147 }{999 \times (1 - 0.761 ) \times 147 + 147 \times (1 - 0.761 ) + 470 } = 4.13
\end{equation}
\begin{equation}
	\text{Speed Up at 90\% sparsity} = \frac{1000 \times 147 }{999 \times (1 - 0.856 ) \times 147 + 147 \times (1 - 0.856 ) + 470 } = 6.79 
\end{equation}

The are additional optimizations that could potentially be applied to further reduce the Top-K overhead which we have not yet evaluated: (1) The overhead reported in Table~\ref{TopKOverhead} is for a Top-K operation, whereas we only need to find a threshold, 
not the top-K weights or activation values and then we can use that threshold for many iterations as suggested by the data in Figure~4 in the paper.  So a more efficient implementation should be possible using the K-Selection Algorithm for which efficient GPU implementations have been proposed~\cite{alabi2012fast}; 
(2) There are more efficient {\em approximate} Top-K algorithms~\cite{chen2011efficient}; (3) Given a slow rate of change in threshold values per iteration, we could potentially {\bf hide} the latency of the Top-K or K-Selection operation during the longer Top-K sampling period by starting the operation earlier.  I.e., we can potentially exploit the stability in thresholds to move obtaining computation of updates to thresholds (after the first iteration) from the critical path; (4) The overhead of finding the Top-K operation on activations in Table~\ref{TopKOverhead} is higher due to large activation size. We speculate this overhead could be reduced significantly by performing the Top-K operation on activation for a single sample and using the resulting threshold for computing the approximate Top-K operation for the entire batch.

\subsection{Top-K Selection} 
\begin{figure}[h]
\centering
\noindent\begin{minipage}{.32\linewidth}
\centering
\includegraphics[width=0.85\textwidth]{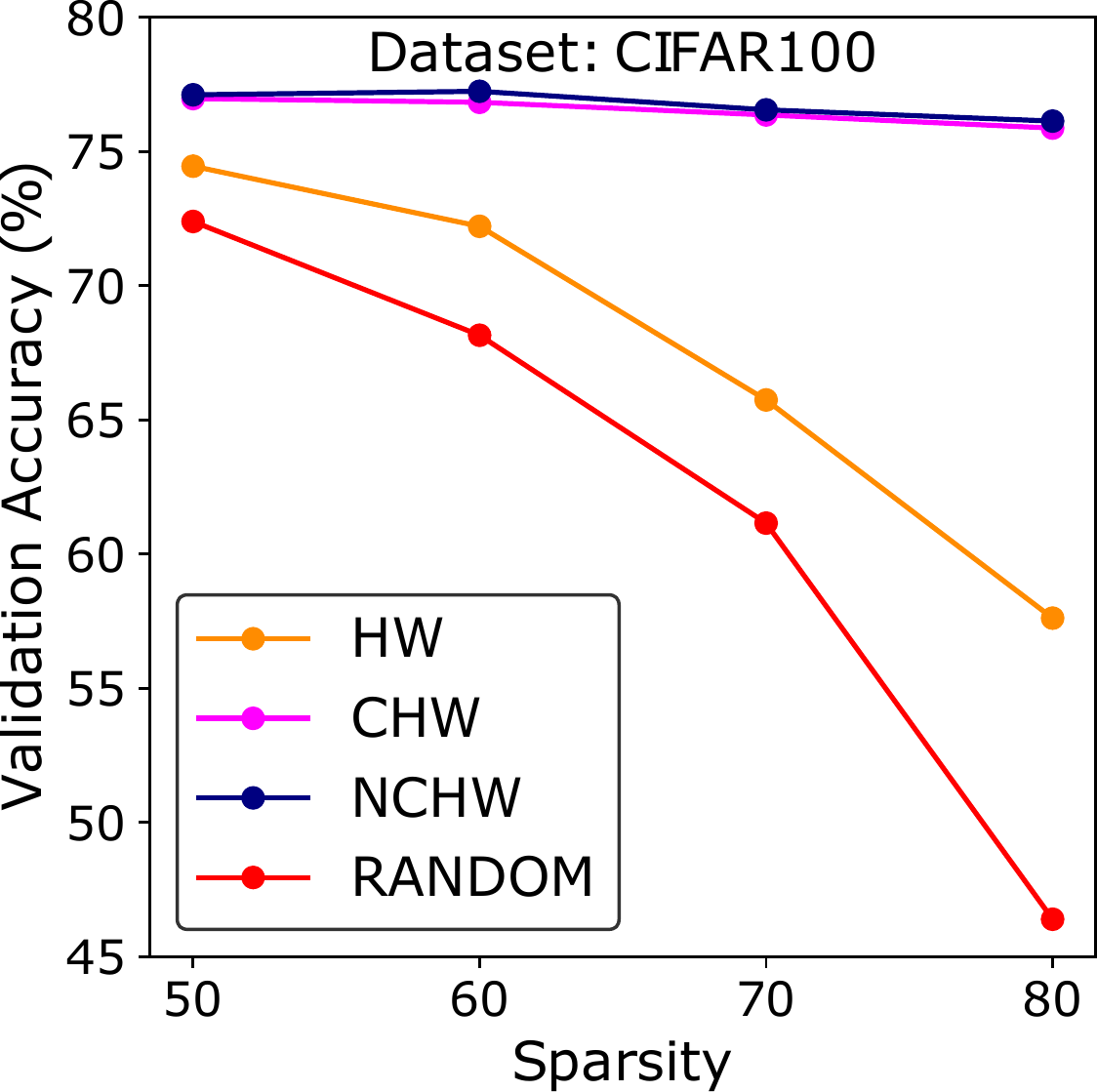}
\end{minipage}%
\begin{minipage}{.67\linewidth}
\centering
\includegraphics[width=\textwidth]{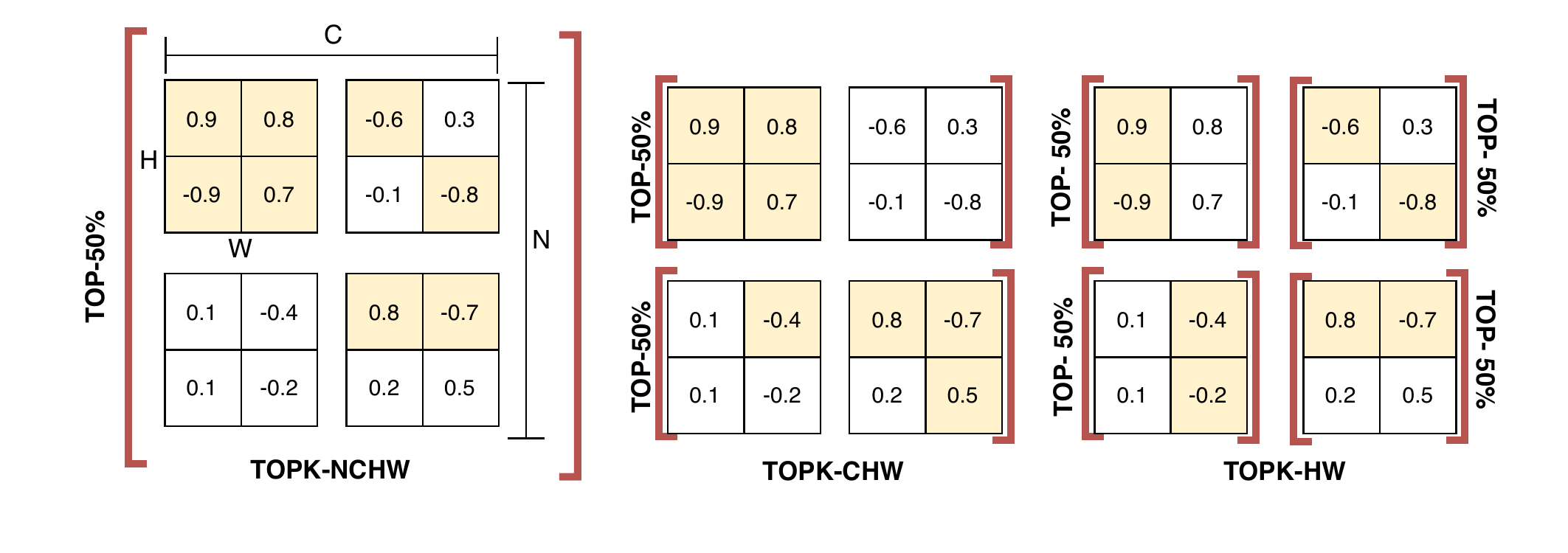}
\end{minipage}
\caption{ \textbf{Different ways of performing top-k operation}. `N' denotes the \#samples in the mini-batch or filters in the layer, `C' denotes the \#channels in the layer. `H' and `W' denote the height and width of the filter/activation map in the layer. Color represent the selected activations/weights by the Top-K operation. }
\label{fig1}
\end{figure}

Given CNNs operate on tensors with many dimensions, there are several options for how to select which components are set to zero during sparsification.
Our CNNs operate on fourth-order tensors, $T~\in~R^{N\times C\times H\times W}$. 
Below we evaluate three variants of the Top-K operation illustrated in the right side of \autoref{fig1}.
We also compared against a null hypothesis in which randomly selected components of a tensor are set to zero.

The first variant, 
labeled TOPK-NCHW in \autoref{fig1},
selects activations and weights to set to zero by considering the entire mini-batch.  
This variant performs Top-K operation over the entire tensor, $f_{TOPK}^{\{N,C,H,W\}}(T)$,
where the superscript represents the dimension along which the Top-K operation is performed.  
The second variant (TOPK-CHW)
performs Top-K operation over the dimensions $C,H$ and $W$ i.e., $f_{TOPK}^{\{C,H,W\}}(T)$ , i.e., selects K \% of input activations from every mini-batch sample and K\% of weights from every filter in the layer. 
The third variant (TOPK-HW) is the strictest form of Top-K operation.  It select K\% of activations or weights from all channels, and thereby performing the Top-K operation over the dimension $H$ and $W$, i.e., $f_{TOPK}^{\{H,W\}}(T_{H,W})$.

The left side of \autoref{fig1} shows the accuracy achieved 
on ResNet-18 for CIFAR100 when using SAW (see Appendix~\ref{AppendixG}) configured with each of these Top-K variants along with a variant where a random subset of components is set to zero.
The results show, first, that randomly selecting works only for low sparsity.  At high sparsity all variants of Top-K outperform random selection by a considerable margin. 
Second, they show that the more constrainted the Top-K operation the less accuracy achieved. 
Constraining Top-K results in selecting some activations or weights which are quite small.  
Similarly, some essential activations and weights are discarded just to satisfy the constraint. 

\subsection{Periodic Top-K} 
\begin{figure}[!t]
\centering
\noindent\begin{minipage}{\linewidth}
\centering
\includegraphics[width=0.7\textwidth]{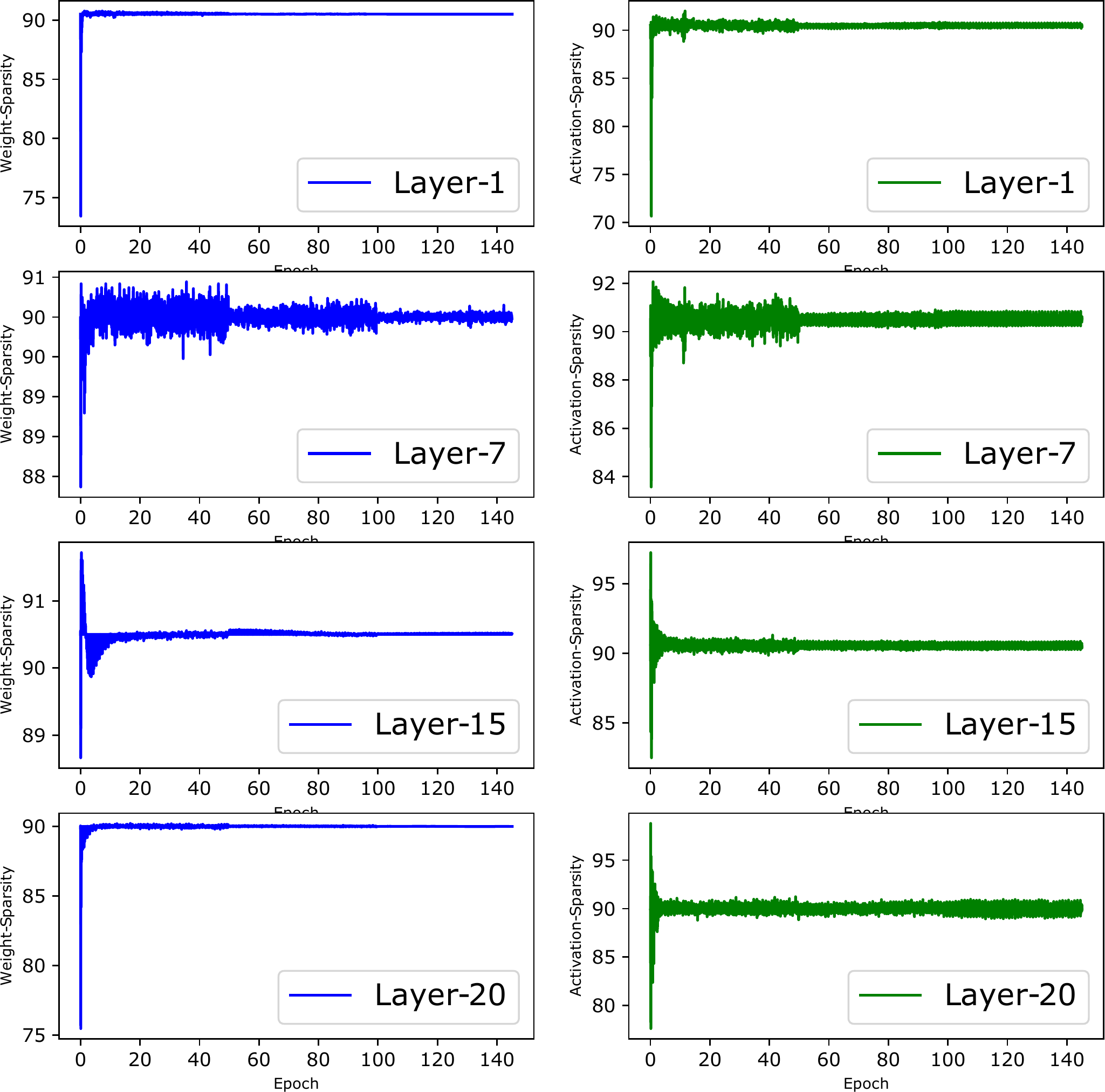}
     \caption{Sparsity Variation using Periodic Top-K Implementation. Network: ResNet-18, Dataset: CIFAR100, Top-K period: 100 iterations, Target Sparsity: 90\%}
     \label{fig:sparsityInPeriodicTopK}
\end{minipage}%
\end{figure}

We have shown there is a little variation in the `K-th' largest element during training, and it remains approximately constant as training proceed. Therefore, the Top-K does not need to be computed every iteration and can be periodically computed after some iterations. We define the number of iterations between computing the threshold for Top-K as the ``Top-K period''.
Since the periodic Top-K used the same threshold during the entire period, therefore, it is crucial to confirm that periodic Top-K implementation does not adversely affect the sparsity during training. We dumped the amount of sparsity obtained in weights and activation using periodic Top-K with period 100 iteration with target sparsity of 90\%.  \autoref{fig:sparsityInPeriodicTopK} shows the sparsity during training using periodic Top-K implementation is concentrated around our targeted sparsity, and the fluctuation decreases as training proceeds confirming our hypothesis that chosen Top-K parameter stabilizes i.e. the Top-K threshold converge to a fixed value during the latter epochs.

\subsection{Sparsification of Output Gradients During Back-Propagation}
\label{AppendixG}
\begin{figure}[t]
     \centering
         \includegraphics[width=0.7\textwidth]{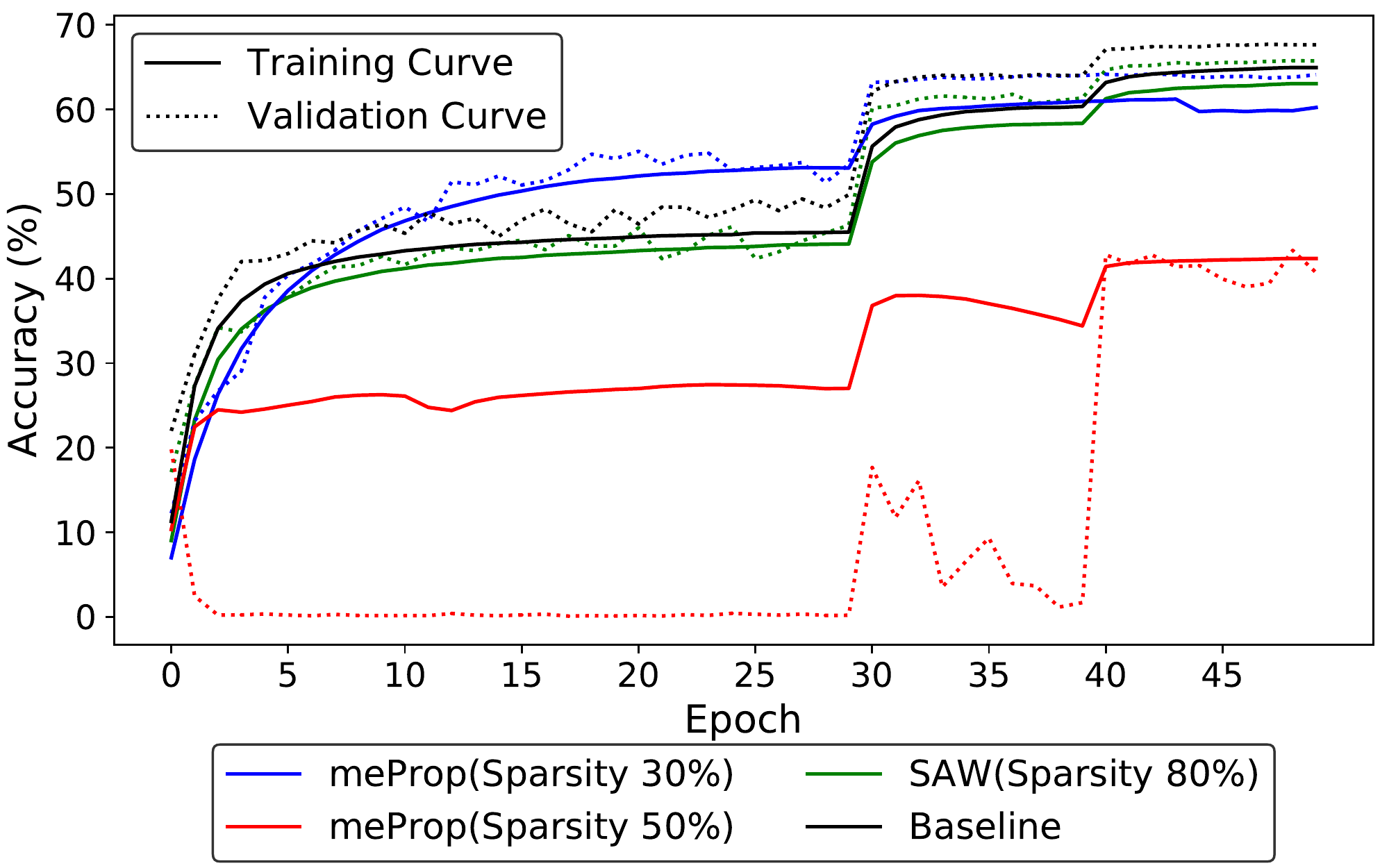}
\caption{\textbf{Convergence Analysis}: Shows the training curve of ResNet18 on ImageNet for meProp and SAW algorithm. Learning rate is reduced by $\frac{1}{10}^{th}$ at $30^{th}$ and $40^{th}$ epoch.}
\label{fig:mePropvsSAW-a}
\end{figure}

SWAT is different from meProp as it uses sparse weight and activation during back-propagation, whereas meProp uses sparse output gradients. 
Our sensitivity analysis shows that convergence is extremely sensitive to the sparsification of output gradients. We compare the performance of the 
meProp and SWAT with deep networks and complex datasets.  To compare SWAT's approach to that of meProp, we use a variant of SWAT-U that only
sparsifies the backward pass; we shall refer to this version of SWAT-U as SAW (Sparse Activation and Weight back-propagation).
\autoref{fig:mePropvsSAW-a} shows SAW and meProp convergence of ResNet18 with the ImageNet dataset. It compares the performance of meProp at 
30\% and 50\% sparsity to SAW at 80\% sparsity. As we can see, meProp converges to a good solution at sparsity of 30\%.  However, at 50\% sparsity, 
meProp suffers from overfitting and fails to generalize (between epochs 5 to 30), and at the same time, it is unable to reach
 an accuracy level above 45\%. These results suggest that dropping output activation gradient ($\bigtriangledown_{a_{l}}$) is generally harmful 
 during back-propagation. On the other hand, SAW succeeds to converge to a higher accuracy even at a sparsity of 80\%. 
 
Moreover, SWAT uses sparse weights and activations in the backward pass allowing compression of weights and activations in the forward pass.
 Effectively,  reducing overall memory access overhead of fetching weights in the backward pass and activations storage overhead because only Top-K\% activations are saved. This memory benefit is not present for meProp since dense weights and activations are needed in the backward pass, whereas there is no storage benefit of sparsifying the output gradients since they are temporary value
 s generated during back-propagation.

\subsection{Sparsification of Batch-Normalization Layer:}
The activations and weights of BN layers are not sparsified in SWAT. Empirically, we found that sparsifying weights and activations are harmful to convergence. This is because the weight (gamma) of BN layers is a scaling factor for an entire output channel, therefore, making even a single BN weight (gamma) zero makes the entire output channel zero. Similarly, dropping activations affects the mean and variance computed by BN. Empirically we found that the BN layer is extremely sensitive to changes in the per channel mean and variance. For example, when ResNet18 is trained on CIFAR 100 using SWAT with 70\% sparsity and we sparsify the BN layer activations, accuracy is degraded by 4.9\% compared to training with SWAT without sparsifying the BN layers. Therefore, the activations of batch-normalization layer are not sparsified. 

The parameters in a BN layer constitute less than 1.01\% to the total parameters in the network and the total computation in the BN layer is less than 0.8\% of the total computation in one forward and backward pass. Therefore, not sparsifying batch-normalization layers only affects the activation overhead in the backward pass.
\newpage

\subsection{Workstation Description}
\begin{center}
\renewcommand{\arraystretch}{1.4} 
\begin{tabular}{|c|c|}
\hline
\hline
\multicolumn{2}{|c|}{WORKSTATION-DESCRIPTION} \\ \hline
\hline
	\multirow{2}{*}{CPU} & Intel(R) Core(TM) i9-9900X CPU @ 3.50GHz\\ 
	          & Intel(R) Xeon(R) Silver 4116 CPU @ 2.10GHz \\ \hline
GPU & NVIDIA 2080-Ti \\ \hline
UBUNTU & Ubuntu 18.04.2 LTS \\ \hline
NVIDIA-DRIVER & 440.33.01 , 418.43 \\ \hline
CUDA, cuDNN & CUDA==10.0.130, cuDNN==7.501 \\ \hline
Pytorch & pytorch==1.1.0, torchvision==0.3.0 \\ \hline
\end{tabular}
\end{center}

\subsection{Details of implementation}
\label{Hyperparameters}
We implemented all models and algorithms on \texttt{pytorch} framework. Code can be found at \url{https://github.com/AamirRaihan/SWAT}. To ease the reproducibility of our experiments, we have also created a docker image.  
We have also uploaded the model checkpoint on anonymous dropbox folder for easily verifying the trained model~\url{https://www.dropbox.com/sh/vo4dxuogk40n6mg/AACdCWWhkhsYdqpjuvsvIb5Oa?dl=0}. 

\begin{table*}[h!]
	\caption{Hyperparameters for ResNet, VGG and DenseNet experiments on CIFAR10/100}
\label{tb:hyperparam}
\centering
\centering
\setlength\tabcolsep{5pt}
\begin{tabular}{l | r r | r r | r r }
  \toprule
    Experiment 
    & \multicolumn{2}{c|}{ 
	\begin{tabular}[t]{@{}c}ResNet~18,~50,~101 \end{tabular} 
    } 
    & \multicolumn{2}{c|}{ 
	\begin{tabular}[t]{@{}c} VGG~16 \end{tabular} 
    }
    & \multicolumn{2}{c}{ 
	\begin{tabular}[t]{@{}c} DenseNet~BC-121 \end{tabular} 
    } \\ \midrule \midrule
  
  Number of training epochs
    & \multicolumn{2}{r|}{150}  
    & \multicolumn{2}{r|}{150}  
    & \multicolumn{2}{r}{150} 
    \\ \midrule
	Mini-batch size ~(\#GPU)      
	& \multicolumn{2}{r|}{128~(1)}  
	& \multicolumn{2}{r|}{128~(1)}  
	& \multicolumn{2}{r}{64~(1)} 
    \\ \midrule  
  \begin{tabular}[t]{@{}l}Learning rate schedule \\ (epoch range: learning rate)\end{tabular} 
    & \begin{tabular}[t]{@{}r@{}@{}@{}}
        1 - 50:\\ 
        51 - 100: \\
        101- 150 : \\
      \end{tabular} 
    & \begin{tabular}[t]{@{}r@{}@{}@{}}
        0.100 \\ 
        0.010 \\
        0.001 \\
      \end{tabular}  
    & \begin{tabular}[t]{@{}r@{}@{}@{}}
        1 - 50:\\ 
        51 - 100: \\
        101- 150 : \\
      \end{tabular} 
    & \begin{tabular}[t]{@{}r@{}@{}@{}}
        0.100 \\ 
        0.010 \\
        0.001 \\
      \end{tabular}  
    & \begin{tabular}[t]{@{}r@{}@{}@{}}
        1 - 50:\\ 
        51 - 100: \\
        101- 150 : \\
      \end{tabular} 
    & \begin{tabular}[t]{@{}r@{}@{}@{}}
        0.100 \\ 
        0.010 \\
        0.001 \\
      \end{tabular}  
    \\  \midrule
  Optimizer
    & \multicolumn{2}{r|}{SGD with Momentum}  
    & \multicolumn{2}{r|}{SGD with Momentum}  
    & \multicolumn{2}{r} {SGD with Momentum} 
    \\ \midrule  
  Momentum
    & \multicolumn{2}{r|}{0.9}  
    & \multicolumn{2}{r|}{0.9}  
    & \multicolumn{2}{r}{0.9} 
    \\ \midrule 
  Nesterov Acceleration
    & \multicolumn{2}{r|}{False}  
    & \multicolumn{2}{r|}{False}  
    & \multicolumn{2}{r}{False}
    \\ \midrule  
  Weight Decay 
    & \multicolumn{2}{r|}{5e-4}  
    & \multicolumn{2}{r|}{5e-4}  
    & \multicolumn{2}{r}{5e-4} 
    \\ \midrule 
  TopK Implementation
    & \multicolumn{2}{r|}{TopK-NCHW}  
    & \multicolumn{2}{r|}{TopK-NCHW}  
    & \multicolumn{2}{r}{TopK-NCHW}
    \\ \midrule 
  TopK Period 
    & \multicolumn{2}{r|}{Per Iteration}  
    & \multicolumn{2}{r|}{Per Iteration}  
    & \multicolumn{2}{r}{Per Iteration}
  \\ \bottomrule

\end{tabular}  
\end{table*}

\begin{table*}[h!]
	\caption{Hyperparameters for WideResNet experiments on CIFAR10/100}
\label{tb:hyperparam}
\centering
\centering
\setlength\tabcolsep{5pt}
\begin{tabular}{l | r r }
  \toprule
    Experiment 
    & \multicolumn{2}{c}{ 
	\begin{tabular}[t]{@{}c}WideResNet~Depth=28~Widen Factor=10 \end{tabular} 
    } \\ \midrule \midrule
  
  Number of training epochs
    & \multicolumn{2}{r}{200}  
    \\ \midrule
	Mini-batch size ~(\#GPU)      
	& \multicolumn{2}{r}{128~(1)}  
    \\ \midrule  
  \begin{tabular}[t]{@{}l}Learning rate schedule \\ (epoch range: learning rate)\end{tabular} 
    & \begin{tabular}[t]{@{}r@{}@{}@{}}
        1 - 60:\\ 
        61 - 120: \\
        121- 160 : \\
        161- 200 : \\
      \end{tabular} 
    & \begin{tabular}[t]{@{}r@{}@{}@{}}
        0.100 \\ 
        0.020 \\
        0.004 \\
        0.0008 \\
      \end{tabular}  
    \\  \midrule
  Optimizer
    & \multicolumn{2}{r}{SGD with Momentum}  
    \\ \midrule  
  Momentum
    & \multicolumn{2}{r}{0.9}  
    \\ \midrule 
  Nesterov Acceleration
    & \multicolumn{2}{r}{True}  
    \\ \midrule  
  Weight Decay 
    & \multicolumn{2}{r}{5e-4}  
    \\ \midrule  
  Dropout Rate 
    & \multicolumn{2}{r}{0.3}  
    \\ \midrule  
  TopK Implementation
    & \multicolumn{2}{r}{TopK-NCHW}  
    \\ \midrule 
  TopK Period 
    & \multicolumn{2}{r}{Per Iteration}  
    \\ \midrule 
  \multirow{3}{*}{Remarks}
    & \multicolumn{2}{c}{\multirow{3}{6.3cm}{First and Last layer are not sparsified since the total parameters in these layers are less than 0.17\% of the total network parameters } }\\
    & \multicolumn{2}{c}{ }\\
    & \multicolumn{2}{c}{ } \\ 
   \bottomrule
\end{tabular}  
\end{table*}

\begin{table*}[h!]
\caption{Hyperparameters for ResNet50/WRN-50-2 experiments on ImageNet}
\label{tb:hyperparam}
\centering
\centering
\setlength\tabcolsep{5pt}
\begin{tabular}{l | r r | r r }
  \toprule
    Experiment 
    & \multicolumn{2}{c|}{ 
	\begin{tabular}[t]{@{}c}SWAT(UnStructured)\end{tabular} 
    } 
    & \multicolumn{2}{c}{ 
	\begin{tabular}[t]{@{}c}SWAT(Structured)\end{tabular} 
    } 
    \\ \midrule \midrule
   Number of training epochs 
    & \multicolumn{2}{r|}{90}  
    & \multicolumn{2}{r}{90}  
    \\ \midrule
   Mini-batch size ~(\#GPU)      
	& \multicolumn{2}{r|}{256~(8)} 
	& \multicolumn{2}{r}{256~(8)}  
    \\ \midrule  
    \begin{tabular}[t]{@{}l}Learning rate schedule \\ (epoch range: learning rate)\end{tabular} 
    & \begin{tabular}[t]{@{}r@{}@{}@{}}
        1 - 30:\\ 
        31 - 60: \\
        61-  80 : \\
        81-  90 : \\
      \end{tabular} 
    & \begin{tabular}[t]{@{}r@{}@{}@{}}
        0.100 \\ 
        0.010 \\
        0.001 \\
        0.0001 \\
      \end{tabular}  
    & \begin{tabular}[t]{@{}r@{}@{}@{}}
        1 - 30:\\ 
        31 - 60: \\
        61-  80 : \\
        81-  90 : \\
      \end{tabular} 
    & \begin{tabular}[t]{@{}r@{}@{}@{}}
        0.100 \\ 
        0.010 \\
        0.001 \\
        0.0001 \\
      \end{tabular}  
    \\ \midrule
	Learning Rate WarmUp      
	& \multicolumn{2}{r|}{Linear~(5 Epochs)}  
	& \multicolumn{2}{r}{Linear~(5 Epochs)}  
    \\  \midrule
  Optimizer
    & \multicolumn{2}{r|}{SGD with Momentum}  
    & \multicolumn{2}{r}{SGD with Momentum}  
    \\ \midrule  
  Momentum
    & \multicolumn{2}{r|}{0.9}  
    & \multicolumn{2}{r}{0.9}  
    \\ \midrule 
  Nesterov Acceleration
    & \multicolumn{2}{r|}{True}  
    & \multicolumn{2}{r}{True}  
    \\ \midrule  
  Weight Decay 
    & \multicolumn{2}{r|}{1e-4}  
    & \multicolumn{2}{r}{1e-4}  
    \\ \midrule 
  Weight Decay on BN parameters 
    & \multicolumn{2}{r|}{No}  
    & \multicolumn{2}{r}{No}  
    \\ \midrule 
  Label Smoothing
    & \multicolumn{2}{r|}{0.1}  
    & \multicolumn{2}{r}{0.1}  
    \\ \midrule 	
  TopK Implementation
    & \multicolumn{2}{r|}{TopK-NCHW}  
    & \multicolumn{2}{r}{TopK-Channel}  
    \\ \midrule 	
  TopK-Period
    & \multicolumn{2}{r|}{1000 iteration}  
    & \multicolumn{2}{r}{1000 iteration}  
    \\ \midrule 	

  \multirow{4}{*}{Remarks}
	& \multicolumn{4}{c}{\multirow{3}{7cm}{First is not sparsified due to low parameters count. To speed up training, we used efficient Top-K implementation where the Top-K is computed periodically after 1000 iteration}} \\
	& \multicolumn{4}{c}{} \\
	\\
  \\ \bottomrule

\end{tabular}  
\end{table*}
\clearpage

\end{appendices}

\end{document}